\documentclass{IEEEtran}
\IEEEoverridecommandlockouts
\usepackage{cite}
\usepackage{amsmath,amssymb,amsfonts}
\usepackage{algorithmic}
\usepackage{graphicx}
\usepackage{textcomp}
\usepackage{xcolor}
\def\BibTeX{{\rm B\kern-.05em{\sc i\kern-.025em b}\kern-.08em
    T\kern-.1667em\lower.7ex\hbox{E}\kern-.125emX}}

\usepackage{tikz}
\usetikzlibrary{patterns}
\usepackage{hyperref}
\usetikzlibrary{automata,positioning,shapes,fit}
\usepackage{mathtools}
\usepackage{amssymb}
\usepackage{algorithm}
\usepackage{xspace}
\usepackage{subfigure}
\usepackage{pgfplots}
\pgfplotsset{compat=1.11}
\pgfplotsset{width=7.5cm,compat=1.12}
\usepgfplotslibrary{fillbetween}
\usepackage{multirow}
\usepackage[switch]{lineno}
\usepackage{makecell}
\usepackage{cleveref}

\newcommand{\numofinputs}{N_{\mathcal I}}
\newcommand{\numofoutputs}{N_{\mathcal O}}
\newcommand{\numoflabels}{h}
\newcommand{\numoffeatures}{m}

\newcommand{\interp}{E}
\newcommand{\interps}{\mathcal E}
\newcommand{\explmeasure}{\Delta_{\interps}}
\newcommand{\corrmeasure}{\Delta_{\mathcal C}}
\newcommand{\blackbox}{\mathcal B}
\newcommand{\prarrow}[1]{\downarrow\!\!#1}
\newcommand{\suarrow}[1]{\uparrow\!\!#1}

\newtheorem{theorem}{Theorem}
\newtheorem{lemma}{Lemma}
\newtheorem{definition}{Definition}
\newtheorem{remark}{Remark}

\newcounter{myctr}

\newenvironment{proof}[0]{\textit{Proof.}}{\hfill$\Box$ \\}

\begin{document}

\title{Synthesizing Pareto-Optimal Interpretations \\
for Black-Box Models
}

\makeatletter
\newcommand{\linebreakand}{%
  \end{@IEEEauthorhalign}
  \hfill\mbox{}\par
  \mbox{}\hfill\begin{@IEEEauthorhalign}
}
\makeatother

\author{\IEEEauthorblockN{Hazem Torfah$^{1}$, Shetal Shah$^{2}$, Supratik Chakraborty$^{2}$, S. Akshay$^{2}$, Sanjit A. Seshia$^{1}$\\}
\IEEEauthorblockA{
\textit{$^{1}$University of California at Berkeley}\\
\{torfah, sseshia\}@berkeley.edu\\
\textit{$^{2}$Indian Institute of Technology, Bombay}
\\
\{shetals, supratik, akshayss\}@cse.iitb.ac.in}
}

\maketitle

\begin{abstract}
We present a new multi-objective optimization approach for synthesizing interpretations that
``explain'' the behavior of black-box machine learning models.
Constructing \emph{human-understandable} interpretations for black-box
models often requires balancing conflicting objectives. A simple
interpretation may be easier to understand for humans while being less
precise in its predictions vis-a-vis a complex
interpretation. Existing methods for synthesizing interpretations use
a single objective function and are often optimized for a single class of
interpretations.  In contrast, we provide a more general and
multi-objective synthesis framework that allows users to choose (1)
the class of syntactic templates from which an interpretation should
be synthesized, and (2) quantitative measures on both the correctness
and explainability of an interpretation. For a given black-box, our
approach yields a set of Pareto-optimal interpretations with respect
to the correctness and explainability measures. We show that the
underlying multi-objective optimization problem can be solved via a
reduction to quantitative constraint solving, such as weighted maximum
satisfiability. To demonstrate the benefits of our approach, we have
applied it to synthesize interpretations for black-box neural-network
classifiers.  Our experiments show that there often exists a rich and
varied set of choices for interpretations that are missed by existing
approaches.

\end{abstract}

\section{Introduction}

Machine learning (ML) components, especially deep neural networks
(DNNs), are increasingly being deployed in domains where
trustworthiness and accountability are major concerns.  Such domains
include health care~\cite{alipanahi2015predicting}, automotive
systems~\cite{NVIDIATegra}, finance~\cite{PayPal}, loans and
mortgages~\cite{sirignano2016deep,forbes2019mortgage}, and
cyber-security~\cite{dahl2013large} among others.  For a system to be
considered accountable and trustworthy, it is necessary to provide
understandable explanations to (possibly expert) humans of why the
system took specific actions/decisions in response to inputs of
concern.  This requires the availability of models that are
human-understandable, and that also predict the outcome of different
components of the system with reasonable accuracy. Laws and
regulations, such as the General Data Protection Regulation (GDPR) in
Europe~\cite{GDPR}, are already emerging with requirements on
explainability of ML components in such systems.  Unfortunately, the
working of ML components like DNNs can be extremely complex to
comprehend, and more so when the components are used as black boxes.
Therefore, there is an urgent need for automated techniques that
generate ``easy-to-understand'' and ``targeted'' interpretations of
black-box ML components, with formal guarantees about the
correctness versus explainability tradeoff.

Synthesizing a ``good'' interpretation of a black-box ML component
often requires striking the right balance between correctness or
accuracy of the interpretation (measured in terms of fidelity,
misclassification rate of predictions etc.) and its explainability or
understandability (approximated by the size/depth of decision
tree/list/diagram, number and nature of predicates used, etc.).  In
most cases, the correctness and explainability measures are in direct
conflict with each other.  Thus, a simple interpretation that is
easily understood by humans may disagree in its predictions with the
output of a black-box ML component for many input instances, whereas
an interpretation that correctly predicts the output for most input
instances may be too large and unwieldy for human comprehension.  This
is not surprising since components like DNNs are often used to learn
highly non-trivial functions for which simple models aren't
available. Therefore, \emph{synthesis of interpretations for black-box
ML components is inherently a multi-objective optimization problem
with conflicting objectives, and Pareto optimality is the best we can
hope for when synthesizing such interpretations.}  %

The literature contains a rich collection of techniques for synthesis
of interpretations for black-box ML components (see, for example,
recent surveys by~\cite{adadi2018peeking} and~\cite{survey-paper}).
Most of these approaches optimize a single correctness measure
(e.g. misclassification rate on a set of samples) while systematically
constraining some explainability measure (e.g. number of nodes or
depth of a decision tree).  Examples of such techniques include
\cite{jha-jar19} wherein sparse logical formulae are synthesized, and
also recent approaches to learning optimal decision trees using
constraint programming~\cite{verwer19,dtip,verhaeghe20},
itemset/rulelist mining~\cite{learnopt-aai20} and SAT-based
techniques~\cite{optimaldt-aaai20,janota20, nina18}, among others.
These approaches often allow efficient generation of a \emph{single}
interpretation with high correctness measure and satisfying
user-provided explainability constraints.  However, no formal
guarantees of Pareto-optimality (w.r.t. correctness and
explainability) are provided. Furthermore, these techniques do not
compute the set of \emph{all} Pareto-optimal interpretations, thereby
constraining the choice of which interpretation to use for a given
application.

In this paper, we present a novel multi-objective optimization
approach for synthesizing Pareto-optimal interpretations of black-box
ML components, using an off-the-shelf quantitative constraint solver
(weighted MaxSAT solver in our case).  For each problem instance, our
approach yields a set of interpretations that correspond to \emph{all}
Pareto-optimal combinations of correctness and explainability
measures.  This contrasts sharply with earlier approaches such
as~\cite{jha-jar19,verwer19,
  dtip,verhaeghe20,learnopt-aai20,optimaldt-aaai20,janota20, nina18}
that always yield a single interpretation, leaving the user with no
choice of exploring the trade-off between correctness and
explainability of alternative interpretations.  Similar to existing
work, we use syntactic constraints to restrict the class of
interpretations over which to search.  Unlike earlier approaches,
however, we do not combine quantitative correctness and explainability
measures into a single optimization objective.  Any such mapping of an
inherently multi-dimensional optimization problem to the
uni-dimensional case results in exclusion of some Pareto-optimal
solutions in general.  Given that quantitative explainability measures
are often just approximations of subjective preferences of the
end-user, we believe it is important to present the entire
set of Pareto-optimal interpretations, and leave the choice of the
``best'' interpretation to the user.  As our experiments show, there is
significant diversity among Pareto-optimal interpretations, and a user
aware of this diversity can make an informed choice for a specific
application.

The syntactic constraints considered in this paper restrict the space
of interpretations to decision diagrams (a generalization of decision
trees) with specified bounds on the number of nodes, predicates and
branching factors.  For simplicity, we let the set of predicates be
pre-determined but potentially large, and with possibly different
relative preferences for different predicates.  We assume that the
black-box ML component model can only be treated as an input-output
oracle, i.e., given an input, we can observe its output and nothing
else. Additionally, we do not have access to training or test data
used to create the black-box component.  Our correctness measure is
therefore based on querying the black-box component with random
samples chosen from its input space, where the sample set size is
carefully chosen to provide statistical guarantees of near-optimality.
Our explainability measure takes into account user preferences of
predicates and also size of the interpretation, prefering smaller
interpretations over larger ones.  The overall framework is, however,
general enough to admit other syntactic classes (beyond decision
diagrams), and also other correctness and explainability measures.

We have implemented our approach in a prototype tool and applied it to
synthesize Pareto-optimal interpretations for some black-box neural
network classifiers.  Our results exhibit the richness of choices
available to the end-user in each case, none of which would be exposed
by existing methods that generate only a single optimal
interpretation.  Indeed, we find that significant improvements in
explainability can sometimes be achieved by only a marginal reduction of
accuracy.

Our primary contributions can be summarized as follows:
\begin{enumerate}
\item We formulate the Pareto-optimal interpretation synthesis problem
  for black-box ML components.
\item We show that finding a single Pareto-optimal interpretation can
  be formulated as a weighted MaxSAT problem, for meaningful
  choices of correctness and explainability scores.
\item We present a divide-and-conquer algorithm for synthesizing
  interpretations for \emph{all} Pareto-optimal
  combinations of correctness and explainability scores.
\item We provide formal guarantees of soundness, completeness and
  universality of our algorithm, and also statistical guarantees of
  near-optimality when only a subset of behaviors of a black-box
  component is sampled.
\item We build a prototype tool and apply it to a collection of
  black-box neural network classifiers: our results show that
  significant diversity exists among Pareto-optimal interpretations
  which earlier tools fail to discover.
\end{enumerate}

\section{Motivating Example}
\label{sec:motivating-example}

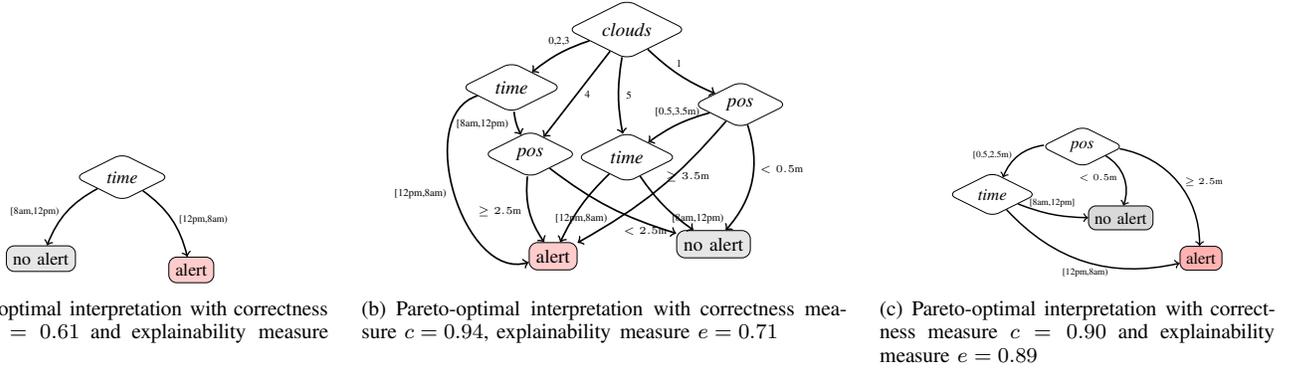
\begin{figure*}[h!]

\subfigure[Pareto-optimal interpretation with correctness measure $c=0.61 $ and explainability measure $e=0.95$]{
\label{fig:bad}
\scalebox{0.75}[0.75]{
\begin{tikzpicture}

\node[] (fake) at (3,0) {\color{white}boo};
\node[] (fake) at (-3,0) {\color{white}boo};

	\node[draw, diamond, rounded corners,aspect=2] (time) at (0,0) {\small \textit{time}};

	\node[draw, rounded corners, below right = 1.2 and 0.4 of time, fill=red!20](alert) {\small alert};
	\node[draw, rounded corners, below left = 1 and 0.4 of time, fill=gray!20](noalert) {\small no alert};

	\path[draw,thick,->] (time) edge[bend left = 20] node[right, align=center]{\tiny [12pm,8am)}(alert);
	\path[draw,thick,->] (time) edge[bend right = 20] node[left]{\tiny [8am,12pm)}(noalert);

\end{tikzpicture}
}
}
\quad
\subfigure[Pareto-optimal interpretation with correctness measure $c=0.94 $, explainability measure $e= 0.71$]{
\label{fig:ugly}
\scalebox{0.8}[0.8]{
\begin{tikzpicture}

\node[] (fake) at (3,0) {\color{white}boo};
\node[] (fake) at (-3,0) {\color{white}boo};

	\node[draw, diamond, rounded corners,aspect=2] (n0) at (0,0) {\small \textit{clouds}};
	\node[draw, diamond, rounded corners,aspect=2, below left = 0.5 and 1 of n0] (n1) {\small \textit{time}};
	\node[draw, diamond, rounded corners,aspect=2, below = 1.2 of n0] (n2) {\small \textit{time}};
	\node[draw, diamond, rounded corners,aspect=2, below right = 0.8 and 1 of n0] (n3) {\small \textit{pos}};
	\node[draw, diamond, rounded corners,aspect=2, below right = 0.7 and -0.5 of n1] (n4) {\small \textit{pos}};
	
	\node[draw, rounded corners, below left = 1.2 and 0.4 of n2, fill=red!20](alert) {\small alert};
	\node[draw, rounded corners, below right = 1 and 0.4 of n2, fill=gray!20](noalert) {\small no alert};
	
	\path[draw,thick,->] (n0) edge[bend right = 10] node[above]{\tiny 0,2,3}(n1);
	\path[draw,thick,->] (n0) edge[bend right = 10] node[right]{\tiny 5}(n2);
	\path[draw,thick,->] (n0) edge[bend right = 0] node[right]{\tiny 4}(n4);
	\path[draw,thick,->] (n0) edge[bend right = 10] node[above]{\tiny 1}(n3);
	
	\path[draw,thick,->] (n1) edge[bend right = 10] node[left]{\tiny [8am,12pm)}(n4);
	\path[draw,thick,->] (n1) edge[bend right = 90] node[left]{\tiny [12pm,8am)}(alert);
	
	\path[draw,thick,->] (n4) edge[bend right = 20] node[left]{\tiny $\ge 2.5$m}(alert);
	\path[draw,thick,->] (n4) edge[bend right = 10] node[below right = 0.2 and 0.1]{\tiny $<2.5$m}(noalert);
	
	\path[draw,thick,->] (n2) edge[bend right = 10] node[below]{\tiny [12pm,8am)}(alert);
	\path[draw,thick,->] (n2) edge[bend right = 10] node[below right]{\tiny [8am,12pm)}(noalert);
	
	\path[draw,thick,->] (n3) edge[bend right = 10] node[above]{\tiny [0.5,3.5m)}(n2);
	\path[draw,thick,->] (n3) edge[bend left = 10] node[above right]{\tiny $\ge 3.5$m }(alert);
	\path[draw,thick,->] (n3) edge[bend left = 30] node[above right]{\tiny $<0.5$m}(noalert);

\end{tikzpicture}
}
}
\quad
\subfigure[Pareto-optimal interpretation with correctness measure $c=0.90 $ and explainability measure $e=0.89 $]{
\label{fig:good}
\scalebox{0.7}[0.7]{
	\begin{tikzpicture}
	
\node[] (fake) at (3,0) {\color{white}boo};
\node[] (fake) at (-3,0) {\color{white}boo};

	\node[draw, diamond, rounded corners,aspect=2] (pos) at (0,0) {\small \textit{pos}};
	\node[draw, diamond, rounded corners,aspect = 2, below left = 0.5 and 0.9 of pos] (time) {\small\textit{time}};
	
	\node[draw, rounded corners, below right = 0 and 1.4 of time, fill=gray!30, ](noalert) {\small no alert};
	\node[draw, rounded corners, below right = 0.3 and 0.5 of noalert, fill=red!30](alert) {\small alert};

	\path[draw,thick,->] (pos) edge[bend left = 40] node[right]{\tiny $\ge 2.5$m}(alert);
	\path[draw,thick,->] (pos) edge[bend left = 40] node[left]{\tiny $<0.5$m}(noalert);
	\path[draw,thick,->] (pos) edge[bend right = 30] node[left]{\tiny [0.5,2.5m)}(time);
	\path[draw,thick,->] (time) edge[bend right = 10] node[ above, align=center]{\tiny [8am,12pm]}(noalert);
	\path[draw,thick,->] (time) edge[bend right = 30] node[ below]{\tiny [12pm,8am)}(alert);
\end{tikzpicture}
}
}

\caption{Pareto-optimal decision diagram interpretations for the black-box monitoring component that decides based on time of day, cloud types, and initial position of an airplane whether to trust a perception module to help the plane track the centerline of a runway. The correctness score is given by the prediction accuracy w.r.t. to the used sample set. The explainability score is the normalized sum of weights of used predicates and  unused nodes.}
\label{fig:example}
\end{figure*} 

We start with an example, adapted from~\cite{fremont-cav20}, that
illustrates the diversity that exists among Pareto-optimal
interpretations of black-box ML models.  Consider a scenario where an
airplane uses a neural network to autonomously taxi along a runway,
relying on a camera sensor. Suppose the plane is expected to follow
the runway centerline within a tolerance of 2.5 meters. The airplane
is equipped with monitoring modules that decide under what
circumstances certain learning-enabled components can be trusted to
behave correctly. One of these monitoring modules
decides under what conditions the camera-based perception module, that
determines the distance to the centerline, can be trusted to deliver
the right values. For example, the monitoring module may use the
weather condition, time of day, and initial positioning of the
airplane to decide whether the perception module's output is reliable.
We wish to reason about this black-box monitoring module, and hence need an
understandable interpretation for it.

Given a set of user-defined predicates (viz. clouds, time of day, and
initial position of the plane), the user may favor certain predicates
over others, and also favor concise interpretations.  By giving
favorability weights to each predicate, we can define an
explainability score that is related to the number of nodes in the
interpretation and also to the predicates used (this is detailed
later). The prediction accuracy of an interpretation is measured w.r.t a set of examples sampled from the black box, and is represented by a
correctness score. Our approach explores the space of interpretations,
searching for concise interpretations that use more favored
predicates and also have high accuracy.  Clearly, to find a ``good''
interpretation that meets these conflicting goals, one must explore
\emph{all} Pareto-optimal interpretations w.r.t. the
criteria above.

\Cref{fig:example} shows three of the many Pareto-optimal
interpretations our approach synthesized for the monitoring black-box.
Each of these has its own pros and cons, and is incomparable with the
others.
The user can now choose the interpretation that best suits the user's 
purpose. For example, if interpretation size is not of concern but
accuracy is, then \Cref{fig:ugly} is the best choice.  However, if the
user wants concise models with favored predicates (related to time of day and initial position), then
\Cref{fig:bad} is the best choice.
The user may also choose the interpretation in \Cref{fig:good}, which
is only slightly less accurate than that in \Cref{fig:ugly}, but has a
higher explainability score. In fact, \Cref{fig:good} represents a
healthy balance between accuracy and explainability.  According to it,
the perception module can be trusted only during morning hours if the
plane starts no more than 2.5m from the centerline, or at any time if
the plane starts within 0.5m of the centerline.

Tools that use a single-objective function to synthesize
interpretations can only find one of these Pareto-optimal
interpretations, depending on the relative weights given to accuracy
and explainability.  The rich diversity among Pareto-optimal
interpretations is completely missed by such tools, effectively
restricting the user's choice of a ``good'' interpretation.

\section{Pareto-optimal Interpretation Synthesis}
\label{sec:po-synthesis}
In this section, we formalize the Pareto-optimal interpretation
synthesis problem and present a solution (for specific choices of
correctness and explainability scores) using a quantitative
constraint satisfaction engine. In our case,
this engine is an off-the-shelf weighted maximum satisfiability
solver.  The key idea is that the user sets syntactic restrictions on
the class of considered interpretations as well as quantitative
objectives for evaluating the interpretations. The quantitative
objectives are defined in terms of two inherently incomparable
measures of interpretations -- the explainability measure and the
correctness measure.  The explainability measure relates to the
``ease'' of understanding of the interpretation by an end-user, while
the correctness measure relates to how precisely the interpretation
explains the behavior of the black-box model on a given set of
samples.  Examples of quantitative correctness measures include
accuracy, recall, precision, F1-score, and many more
\cite{introToDataMining}. Examples of explainability measures include
those that reward usage of concise interpretations and less complex
predicates, among others.

Since our access to the black-box model is only via input/output
samples, the correctness measure referred to above is defined with
respect to a set of samples, and not with respect to the black-box
model in its entirety.  While this may appear ad-hoc at
first sight, we show in \Cref{sec:refinement} that rigorous
statistical guarantees can indeed be provided with sufficiently many
samples.

\subsection{Formal problem definition}

We now give a formal definition of the Pareto-optimal interpretation
synthesis problem.
An interpretation is simply a syntactic structure, viz. decision tree,
decision diagram, linear model, etc. We will fix a class of
interpretations $\interps$ over an input domain $\mathcal I$ and
output domain $\mathcal O$. For an interpretation $\interp \in
\interps$, we define $f_{\interp}\in (\mathcal I \rightarrow \mathcal
O)$ to be the semantic function that is computed by~$\interp$. Note
that different interpretations may compute the same semantic function.

Every interpretation $E \in \interps$ is associated with a pair of
real-valued measures $(c,e)$, where $c$ is the correctness measure and $e$ is the
explainability measure of $E$.
We define a partial order $\preceq$ on such pairs as: $(c,e)\preceq
(c',e')$ iff $c\leq c'$ and $e\leq e'$. Given a set $X$ of $(c,e)$
pairs, we define $\max^{\preceq} ~X$ to be the set of
$\preceq$-maximal pairs in $X$. An interpretation $E$ with the pair of measures
$(c,e)$ is said to be \emph{Pareto-optimal} if $(c,e)$ is maximal
over pairs of measures of all interpretations.
\begin{definition}[Pareto-optimal interpretation synthesis]
  \label{def:qsygus}
	Let $\interps$ be a syntactic class of interpretations over
        inputs $\mathcal I$ and outputs $\mathcal O$. Further, let
        $\mathcal S \subseteq \mathcal I \times \mathcal O$ be a set
        of samples, $\corrmeasure \colon(\mathcal I \rightarrow
        \mathcal O)\times 2^{(\mathcal I \times \mathcal
          O)}\rightarrow {\mathbb R}^{\ge 0}$ be a correctness
        measure, and $\explmeasure \colon \mathcal E \rightarrow
        {\mathbb R}^{\ge 0}$ an explainability measure.  The
        Pareto-optimal interpretation synthesis problem $\langle
        \mathcal E, \mathcal S, \corrmeasure, \explmeasure \rangle$ is
        the multi-objective problem of finding a Pareto-optimal
        interpretation $\interp \in {\arg\max^\preceq_{\interp'
          \in~\interps}}~(\corrmeasure(f_{\interp'},\mathcal
        S), \explmeasure(E'))$.  
\end{definition}

 We interpret $\corrmeasure(f_\interp, \mathcal S)$ as a measure of
 closeness between the semantic function $f_\interp$ of interpretation
 $\interp$ and the semantic constraints defined by a set $\mathcal S$
 of samples. An optimally correct interpretation is one with maximal
 closeness. An example of such a measure is the \emph{prediction
 accuracy} %
 $\frac{|\{(i,o)\in \mathcal S \mid f_{\interp}(i) = o\}|}{|\mathcal
   S|}$. The problem can also be defined in terms of the ``distance''
 between an interpretation and the semantic constraints defined by
 $\mathcal S$, in which case, the optimization problem is one of
 minimization.  An example of such a measure is the
 \emph{misclassification rate}, which is one minus the prediction
 accuracy. Similarly, for $\explmeasure(\cdot)$, we choose to define
 it as a reward function that we want to maximize, but it can also be
 dually defined as a cost function we want to minimize.

 For each $\preceq$-maximal pair of measures, there can be multiple
 corresponding interpretations realizing the measures.  We don't
 distinguish between them for purposes of this paper. The following
 definition is therefore relevant.
 \begin{definition}[Minimal representative set]
\label{def:minrepset}   A set
 $\Gamma$ of Pareto-optimal interpretations is a
  minimal representative set for $\langle \mathcal E, \mathcal S,
  \corrmeasure, \explmeasure \rangle$ if for every $(c, e) \in
  {\max^\preceq_{\interp \in \interps}}(\corrmeasure(f_{\interp},\mathcal S),
  \explmeasure(E))$, there is exactly one interpretation $\interp' \in
  \Gamma$ such that $(\corrmeasure(f_{\interp'},\mathcal S),
  \explmeasure(\interp')) = (c, e)$.
 \end{definition}
 Our goal can therefore be stated as one of finding a minimal
 representative set of interpretations for a black-box model.

\subsection{Synthesis via weighted maximum satisfiability}
\label{sec:maxsat-enc}
We now discuss how to synthesize one (of possibly many) Pareto-optimal
interpretation for specific choices of $\interps$, $\corrmeasure$ and
$\explmeasure$, by encoding the synthesis problem as a \emph{weighted
maximum satisfiability} problem (weighted \textsc{MaxSat}).  For
purposes of our discussion, we choose $\interps$ to be the class of
\emph{bounded multi-valued decision diagrams}, i.e., decision diagrams
with multiple branching at each node, where the branching is governed
by decision predicates, and with a bound on the number of decision
nodes (see, for example, diamond nodes in \Cref{fig:example}).  We use
prediction accuracy as the correctness measure, and define the
explainability measure with weights (denoting preferences) on the
predicates and on the number of used nodes.  The encoding for several
other classes of interpretations, such as decision trees, decision
rules, etc. and for other explainability and correctness measures can
be done similarly.

We start with a brief recap of the weighted \textsc{MaxSAT} problem. A
Boolean formula $\varphi$ over variables in a set $X$ is said to be in
conjunctive normal form (CNF) if $\varphi$ is of the form $C_1 \wedge
C_2 \wedge \cdots C_m$, where each $C_i$ is a disjunction of literals
(i.e. variables or negations of variables).  An assignment
$\sigma\colon X \rightarrow \{0,1\}$ is an assignment of truth values
to variables. If a clause $C_i$ evaluates to $1$ under $\sigma$, we
say $\sigma$ satisfies $C_i$, denoted by $\sigma \models C_i$.

\begin{definition}[{Weighted Maximum Satisfiability}]
\label{def:maxsat}
Given a Boolean formula $\varphi = \bigwedge_{i=1}^m C_i$ in CNF and a weight
function $w\colon\{C_1, \ldots C_m\} \rightarrow \mathbb R^{\ge 0}$ that assigns a non-negative real weight to each clause,
the weighted \textsc{MaxSAT} problem asks us to find an assignment $\sigma$ such that $\sum_{\{C_i \mid ~ \sigma \models C_i\}} w(C_i)$ is maximized.
\end{definition}
In a variant of the above definition, the clauses in $\varphi$ are partitioned into \emph{hard} and \emph{soft} clauses. The problem now is to find an assignment $\sigma$ that satisfies \emph{all hard clauses} and maximizes the sum of weights of satisfied soft clauses.  We use this variant for encoding our problem.%

 At a high level, for an instance $\langle \interps, \mathcal S,
 \corrmeasure, \explmeasure \rangle$ of the Pareto-optimal
 interpretation synthesis problem, the encoding is defined as a
 conjunction of four formulae.  Specifically, $\phi_{\langle \interps,
   \mathcal S, \corrmeasure, \explmeasure \rangle} = \phi_{\interps}
 \wedge \phi_{\mathcal S} \wedge \phi_{\corrmeasure} \wedge
 \phi_{\explmeasure}$, where $\phi_{\interps}$ encodes the syntactic
 restrictions, i.e., bounded multi-valued decision diagrams with the
 permitted predicates (features and branchings) and labels, and
 $\phi_{\mathcal S}$ encodes the semantic constraints, i.e., the
 relation between the samples in $\mathcal S$ and an interpretation
 satisfying $\phi_{\interps}$. The formula $\phi_{\corrmeasure}$
 encodes the correctness measure, e.g., in case of prediction
 accuracy, it encodes whether an interpretation agrees on a
 sample. Lastly, $\phi_{\explmeasure}$ defines constraints that encode
 certain structural aspects of an interpretation, e.g., what
 predicates were chosen and whether a node was used. We discuss some
 details of these formulas below, leaving the full encoding to
 the Appendix.

\paragraph{Encoding of the interpretation class ($\phi_{\interps}$)}
We discuss the encoding for bounded multi-valued decision diagrams over inputs $\mathcal I$ and outputs $\mathcal O$. The diagrams are restricted by a finite set of decision predicates, denoted by $P$.
For example, in \Cref{fig:bad}, the initial node uses the ``{\it time of day}'' predicate with branchings: \{[8am-12pm], [12pm-8am]\}. %
Let $L$
be a set of output labels. In \Cref{fig:example}, we have two labels, ``\emph{alert}'' and ``\emph{no alert}''.  
An \emph{interpretation} $\interp \in \interps$ is a multi-valued decision diagram over a finite set of nodes $\mathcal N$, where each internal node corresponds to a decision predicate $p\in P$ and each leaf to  an output label $\ell \in L$.  Outgoing transitions of a node are labelled according to the branchings of the predicate corresponding to the node. 
We remark that features are distinct from inputs to the black-box. For example, in the decision diagrams in \Cref{fig:example} the feature ``\emph{pos}'' uses the latitude and longitude inputs to compute the initial position of the plane. 
Furthermore, the same predicate may appear on different nodes in the decision diagram, but not more than once along a path.
For a given $P$, $L$, and a bound $n$ on the number of nodes $\mathcal N$ in the decision diagram, the formula $\phi_{\interps}$ encodes an acyclic decision diagram of at most $n$-nodes over a set $P$ of predicates, with leaves labeled by elements of $L$. 

\paragraph{Encoding of the samples} The formula $\phi_{\mathcal S}$ encodes the relation between the samples and the interpretation $\phi_{\interps}$. 
It uses an auxiliary variable $m_{(i,o)}$ for each sample $(i, o)$ in
the set ${\mathcal S}$. Logically, $m_{(i,o)}$ is set to true iff the
interpretation given by a satisfying assignment of $\phi_{\interps}$
produces the output label $o$ when fed the input $i$.  For decision
diagrams, this is encoded by symbolically matching the input $i$ to a
decision path in the diagram, and by comparing the value of $o$ with
that of the label reached at the end of the decision path.  Note that
the count of these auxiliary variables grows linearly with the size of
the sample set.

\paragraph{Encoding the correctness measure ($\phi_{\corrmeasure}$)}
To encode $\corrmeasure$, we add a unit soft clause (i.e., a clause
with only one literal) $m_{(i,o)}$ for each sample $(i, o)$.%
By assigning appropriate weights to these unit clauses and by
maximizing the sum of weights of satisfied clauses (see
\Cref{def:maxsat}), we obtain an interpretation that maximizes
$\corrmeasure$ with respect to the sample set ${\mathcal S}$.  E.g.,
if $\corrmeasure$ represents the prediction accuracy, then assigning a
weight of $1$ to each unit clause $m_{(i,o)}$ gives us an
interpretation that agrees on a maximal number of samples in $\mathcal
S$. If the user is interested in interpretations that agree on certain
types of samples, then higher weights should be given to these
samples.  Explicitly, to define such measures $\corrmeasure$, the user
can provide a function $w\colon \mathcal I \times \mathcal O
\rightarrow \mathbb R$, that defines these weights. For example, in
the case of prediction accuracy, $w$ is the constant function~1.

\paragraph{Encoding the explainability measure ($\phi_{\explmeasure}$)}
To encode $\explmeasure$, we add a unit clause $u_\gamma$ for each
syntactic structure $\gamma$ of an interpretation in $\interps$ and
give it a weight according to how favorable $\gamma$ is. For example,
in the case of decision diagrams, using some predicates may be more
favorable than others.  To encode this, we add unit clauses
$u_{(i,p)}$ that are set to true iff predicate $p$ is used in  node
$i$, and assign higher weights for clauses representing favorable
predicates.
Moreover, predicates with fewer branches can be favored by using soft clauses with appropriate weights. %
To further reward the synthesis of decision diagrams with fewer nodes, we can also add unit soft clauses $u_i$ for each node $i$ that is set to true iff node $i$ is not reachable from the root node in an interpretation satisfying $\phi_{\interps}$, and give them positive weights. In this case, by maximizing the satisfaction of these clauses, we reward the synthesis of small decision diagrams.

In our weighted \textsc{MaxSAT} formulation, we require that all clauses resulting from a Tseitin encoding (i.e., a transformation into CNF) of formula $\phi_{\langle \interps,\mathcal S, \corrmeasure, \explmeasure \rangle }$, except for unit soft clauses mentioned above, be hard clauses.  
On feeding the above to a \textsc{MaxSAT} solver, it returns a satisfying assignment giving a concrete instantiation of the decision diagram template that maximizes the sum of weights of $m_{(i,o)}$ and $u_{\gamma}$ clauses.

Under the assumption that the class of interpretations and
explainability and correctness measures are encodable as Boolean
formulas, we have the following theorem. Note that this assumption is
not necessarily restrictive in practice. For most types of
interpretation classes used in the literature, viz. decision trees,
decision diagrams, decision lists and sets of bounded depth/size, and
also for measures such as accuracy with its many weighted variants,
the problem is indeed encodable as a weighted \textsc{MaxSAT}
instance.
\begin{theorem}[Pareto-optimality]
  \label{thm:pareto-opt}
  Every solution of the weighted \textsc{MaxSAT} problem
  $\phi_{\langle \interps,\mathcal S, \corrmeasure, \explmeasure
    \rangle }$ gives a solution for the Pareto-optimal interpretation
  synthesis problem ${\langle \interps,\mathcal S, \corrmeasure,
    \explmeasure \rangle }$.
\end{theorem}

\subsection{Exploring the set of Pareto-optimal interpretations}
\label{sec:exploring-po-int}

\begin{figure*}[t]
\scalebox{0.75}[0.75]{
\subfigure[First iteration: Exploring region defined by bounds $(0,1,0)$. Expand $W$ with new regions $R_3^{c,e}$ and $R_4^{c,e}$ by adding the points $(0,\prarrow{e_0},c_0)$ and $(\suarrow{e_0},1,0)$. No Pareto-optimal points exist in the red region.]{\label{fig:iter1}
\begin{tikzpicture}
	\begin{axis}[xmin=0, xmax=1, ymin=0, ymax=1, samples=50,xlabel={$\explmeasure$},
    ylabel={$\corrmeasure$}]
    \addplot[blue, mark=square*] coordinates {(0.7,0.6)} node[below left] {\small $(c,e)$};
    \draw[pattern=dots, pattern color=green, draw=green] (0.7,0.6) -- (0,0.6);
    \draw[pattern=dots, pattern color=green, draw=green] (0.7,0.6) -- (0.7,0);
    \draw[pattern=north west lines, pattern color=red, draw=red] (0.7,0.6) rectangle (1,1);
    \node[] at (0.85,0.8){$R_1^{c,e}$};
    \node[] at (0.37,0.3){$R_2^{c,e}$};
    \node[] at (0.37,0.8){$R_3^{c,e}$};
    \node[] at (0.85,0.3){$R_4^{c,e}$};
    
	\end{axis}
	
\end{tikzpicture}
}
}
\scalebox{0.75}[0.75]{
\subfigure[Exploring the region $R_3^{c,e}$. A new Pareto-optimal interpretation is found with measures $(c',e')$. Add the points $(0,\prarrow{e'},c')$ and $(\suarrow{e'},\prarrow{e},c)$ to $W$.]{\label{fig:iter2}
\begin{tikzpicture}
	\begin{axis}[xmin=0, xmax=1, ymin=0, ymax=1, samples=50,xlabel={$\explmeasure$},
    ylabel={$\corrmeasure$}]
    \addplot[blue, mark=square*] coordinates {(0.7,0.6)} node[below left] {\small $(c,e)$};
    \draw[pattern=dots, pattern color=green, draw=green] (0.7,0.6) -- (0,0.6);
    \draw[pattern=dots, pattern color=green, draw=green] (0.7,0.6) -- (0.7,0);
    \draw[pattern=north west lines, pattern color=red, draw=red] (0.7,0.6) rectangle (1,1);
    \node[] at (0.85,0.3){$R_4^{c,e}$};

	\addplot[blue, mark=square*] coordinates {(0.2,0.85)} node[below left] {\small $(c',e')$};
	\draw[pattern=dots, pattern color=green, draw=green] (0.2,0.85) -- (0,0.85);
    \draw[pattern=dots, pattern color=green, draw=green] (0.2,0.85) -- (0.2,0);
    \draw[pattern=north west lines, pattern color=red, draw=red] (0.2,0.85) rectangle (0.7,1);
    \node[] at (0.45,0.7){$R_4^{c',e'}$};
    \node[] at (0.1,0.92){$R_3^{c',e'}$};
    \node[] at (0.45,0.92){$R_1^{c',e'}$};

	\end{axis}
	
\end{tikzpicture}
}
}
\scalebox{0.75}[0.75]{
\subfigure[Exploring region $R_4^{c',e'}$. Optimal interpretation had correctness measure $c''< c$. Exclude region $R_1^{c,e''}$ and add new region defined by $(\suarrow{e'},\prarrow{e''},c)$ to $W$. For another Pareto-optimal point $(c''',e''')$, no solution found when exploring its region $R_3^{c''',e'''}$. ]{\label{fig:iter3}
\begin{tikzpicture}
	\begin{axis}[xmin=0, xmax=1, ymin=0, ymax=1, samples=50,xlabel={$\explmeasure$},
    ylabel={$\corrmeasure$}]
    \addplot[blue, mark=square*] coordinates {(0.7,0.6)} node[below left] {\small $(c,e)$};
    \draw[pattern=dots, pattern color=green, draw=green] (0.7,0.6) -- (0,0.6);
    \draw[pattern=dots, pattern color=green, draw=green] (0.7,0.6) -- (0.7,0);
    \draw[pattern=north west lines, pattern color=red, draw=red] (0.7,0.6) rectangle (1,1);
    
    \addplot[blue, mark=square*] coordinates {(0.2,0.85)} node[below left] {\small $(c',e')$};
	\draw[pattern=dots, pattern color=green, draw=green] (0.2,0.85) -- (0,0.85);
    \draw[pattern=dots, pattern color=green, draw=green] (0.2,0.85) -- (0.2,0);
    \draw[pattern=north west lines, pattern color=red, draw=red] (0.2,0.85) rectangle (0.7,1);
    \node[] at (0.1,0.92){$R_3^{c',e'}$};
    \node[] at (0.45,0.92){$R_1^{c',e'}$};

    \addplot[orange, mark=square*] coordinates {(0.5,0.3)} node[below left] {\small $(c'',e'')$};
    \addplot[orange, mark=square*] coordinates {(0.5,0.6)} node[below left] {\small $(c,e'')$};
    \draw[pattern=dots, pattern color=green, draw=orange] (0.5,0) -- (0.5,.85);
    \draw[pattern=north west lines, pattern color=red, draw=red] (0.5,0.6) rectangle (0.7,0.85);
   	\node[] at (0.6,0.75){$R_1^{c,e''}$};
   	\node[] at (0.35,0.75){$R_3^{c,e''}$};
   	
   	\addplot[blue, mark=square*] coordinates {(0.9,0.3)} node[below] {\small $(c''',e''')$};
   	\draw[pattern=north west lines, pattern color=red, draw=red] (0.7,0.6) rectangle (0.9,0.3);
   	\node[] at (0.8,0.45){\small $R_3^{c''',e'''}$};

	\end{axis}
	
\end{tikzpicture}
}
}

\caption{An illustration of \Cref{alg:allintsynt}.}
\label{fig:illustration_AllIntSynt}
\end{figure*}
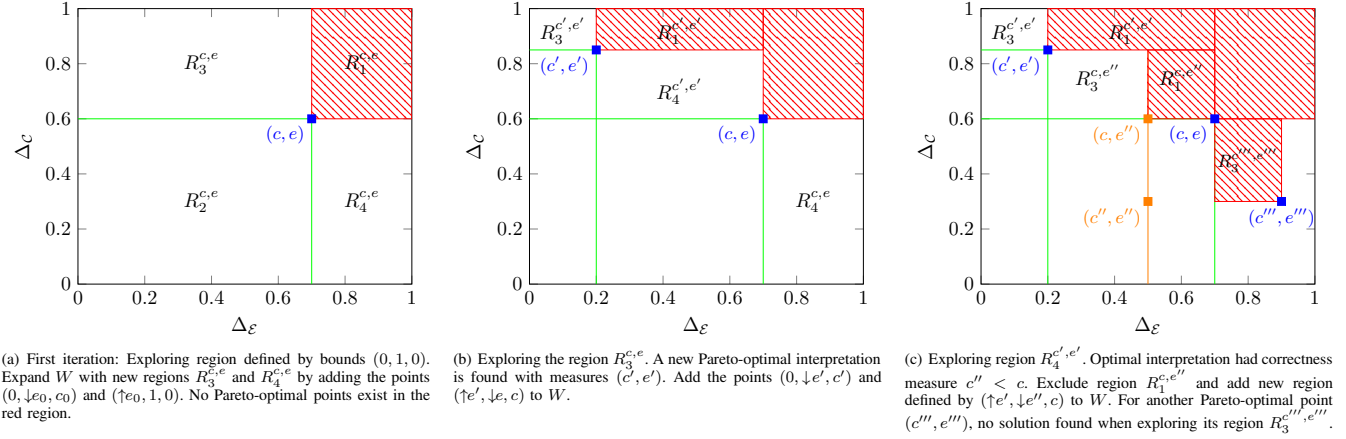

We now present an algorithm for computing a minimal representative set
of Pareto-optimal interpretations.
The algorithm is based on the key observation that every Pareto-optimal measure $(c,e)$ splits the space of measures into four regions, depicted in \Cref{fig:iter1}, (1) a region $R_1^{c,e}$ of measures for which there exists no solution, namely, all measures $(c',e')\not =(c,e)$ with $c'\ge c$ and $e'\ge e$, otherwise $(c,e)$ would not be Pareto-optimal, (2) a region $R_2^{c,e}$ of measures that are not Pareto-optimal, namely, all points $(c',e')\not =(c,e)$ with $c'\leq c$ and $e'\leq e$, (3) a region $R_3^{c,e}$ with measures of potential Pareto-optimal interpretations with better correctness measures, i.e., those with measures $(c',e')$ with $c'>c$ and $e'<e$, and lastly (4) a region $R_4^{c,e}$ with measures of potential Pareto-optimal interpretations with better explainability measures, i.e., points $(c',e')$ with $c'<c$ and $e'>e$. By synthesizing a first Pareto-optimal interpretation using the procedure from last section, and then dividing the search space into the corresponding regions (1)-(4), our algorithm proceeds by searching for further Pareto-optimal interpretations with better correctness in region (3) and better explainability in region (4). This process is repeated for every Pareto-optimal interpretation found by our algorithm, thus, directing the search into smaller and smaller regions until no new Pareto-optimal interpretation can be found. 

The algorithm is given in \Cref{alg:allintsynt} and the exploration process it implements is illustrated in \Cref{fig:illustration_AllIntSynt}. For $\interps, \mathcal S, \corrmeasure$, and  $\explmeasure$, \Cref{alg:allintsynt} returns a
minimal representative set $\Gamma$ of interpretations for all Pareto-optimal measures.
To synthesize a Pareto-optimal interpretation within a given region of
measures, \Cref{alg:allintsynt} relies on the procedure
\textsc{QuIntSynt} which given $\interps, \mathcal S, \corrmeasure$,
and $\explmeasure$, in addition to a lower-bound $\delta_\interps^l$
and upper-bound $\delta_\interps^u$ on the explainability measure,
returns a Pareto-optimal interpretation $\interp$ with explainability
measure $e$ such that $\delta_\interps^l \leq e \leq
\delta_\interps^u$.  \textsc{QuIntSynt} effectively solves an
extension of the weighted MaxSAT instance defined in the last section,
in which we additionally require the explainability measure to satisfy
the constraints given by the lower-bound $\delta_\interps^l$ and
upper-bound $\delta_\interps^u$.  This can be done by extending the
formula $\phi$ in the last section with a fifth conjunct
$\phi_{\delta_\interps^l,\delta_\interps^u}$. This conjunct is
satisfied if the sum of weights of the used syntactic structures
(e.g. in the case of decision diagrams, this will be sum of weights of
the satisfied clauses $u_{(i,p)}$ and ${u_i}$) lies within the given
bounds. We leave details of this encoding to the Appendix, but
intuitively, we encode a binary adder that sums up the weights of
satisfied $u_{(i,p)}$ and ${u_i}$ clauses and compare the results to
binary encodings of the bounds.  To fix the number of bits to encode
both the adder and bounds, we normalize the weights to values between
$0$ and $1$ up to a certain floating-point precision $k$.  Next we
explain \Cref{alg:allintsynt} in some detail, elaborating on
why it suffices to only bound the explainability measure when exploring
regions (3) and (4) depicted in \Cref{fig:iter1}.

\begin{algorithm}[t]
\caption{\textsc{ExplorePOI}}
\begin{algorithmic}[1]
	\REQUIRE $\interps$, $\mathcal S$, $\corrmeasure$, $\explmeasure$
	\ENSURE Minimal representative set $\Gamma$ for  $\langle \interps$, $\mathcal S$, $\corrmeasure$, $\explmeasure\rangle$   \vspace{0.2cm}
	\STATE $\Gamma := \emptyset$
	\STATE $W := \{(0,1,0)\}$
	\WHILE {$W \not = \emptyset$}
		\STATE $(\delta_\interps^l, \delta_\interps^u,\delta_{\mathcal C})$ := pop($W$)
		\STATE $(E,(c,e)) = $ \textsc{QuIntSynt}$(\interps,\mathcal S, \corrmeasure, \explmeasure, \delta_\interps^l, \delta_\interps^u)$
		
		\IF {$E \not =\bot$}
			\IF{$c > \delta_{\mathcal C}$}
				\STATE $\Gamma := \Gamma \cup \{(E,(c,e)\}$
				\STATE push$(W,(\delta_\interps^l, \prarrow{e}, c))$
				\STATE push$(W, (\suarrow{e},  \delta_{\interps}^u, \delta_{\mathcal C}))$
			\ELSE
				\STATE push$(W,(\delta_\interps^l, \prarrow{e},\delta_{\mathcal C}))$
			\ENDIF
		\ENDIF
	\ENDWHILE
	\RETURN $\Gamma$
	
\end{algorithmic}
\label{alg:allintsynt}
\end{algorithm}

Initially, \Cref{alg:allintsynt} explores the entire set of Pareto-optimal solution space. To this end, the exploration set $W$ is initialized with the point $(0,1,0)$ (line 2) defining a lower bound on the explainability measure, an upper-bound on the explainability measure, and a lower-bound on the correctness measure, respectively. For every point $(\delta_\interps^l, \delta_\interps^u,\delta_{\mathcal C})$ in $W$, \textsc{QuIntSynt} synthesizes a Pareto-optimal region within the explainability measure bounds defined by $\delta_\interps^l$ and $ \delta_\interps^u$ (line 5). If an interpretation $\interp$ is found with measures $c$ and $e$, i.e., $\interp \not =\bot$ (line 6), the algorithm further divides the search space based on the following case distinction: 
\begin{itemize}
\item if $c>\delta_{\mathcal C}$, then a new Pareto-optimal interpretation with measures $(c,e)$ is found and the regions $R_3^{c,e}$ and $R_4^{c,e}$ defined by the points $(\delta_\interps^l, \prarrow{e},c)$ and $(\suarrow{e} ,\delta_\interps^u,\delta_{\mathcal C})$, respectively, are added to $W$ (lines 9 and 10). The operators $\prarrow{ }$ and $\suarrow{ }$ define the  predecessor and successor value of the value $e$ (we assume that the values are discrete and hence the predecessor and successor exist).
	For example, if the interpretation synthesized by \textsc{QuIntSynt} is one with measures $c',e'$ as depicted in \Cref{fig:iter2}, then the region $R_4^{c',e'}$ is be captured by the point $(\suarrow(e'), \prarrow(e), c)$. The region $R_3^{c',e'}$ is captured by $(0,\prarrow(e'),c')$. 	Notice that we do not need to include an upper bound on the correctness measure as it is already implicitly defined by the $R_1^{c,e}$ region of any Pareto-optimal point $(c,e)$. For example, in \Cref{fig:iter2} the upper bound on the correctness for region $R_4^{c',e'}$ is already captured through the fact that no Pareto-optimal solutions exist in $R_1^{c',e'}$.     
	\item if $c \leq \delta_{\mathcal C}$, then $(c,e)$ cannot be Pareto-optimal, because we already know that there is a Pareto-optimal interpretation with measures $(\delta_{\mathcal C}, \suarrow{\delta_\interps^u})$. In this case, we can exclude the search in the region $R_1^{\delta_{\mathcal C},e}$, because if there was any Pareto-optimal interpretation with measures $(\hat{c},\hat{e})$ in $R_1^{\delta_{\mathcal C},e}$, then \textsc{QuIntSynt} would have found this interpretation. Thus, \Cref{alg:allintsynt} further prunes the search region to a smaller region defined by $(\delta_\interps^l,\prarrow{e},\delta_{\mathcal C})$ (line 12).  
	For example, if \Cref{alg:allintsynt} used \textsc{QuIntSynt} to synthesize an interpretation from $R_4^{c',e'}$, and returned a solution with measures $(c'',e'')$ as depicted in \Cref{fig:iter3}, then we can exclude the search in region $R_1^{c,e''}$ and add the region $R_3^{c,e''}$ to $W$.   
\end{itemize}

Lastly, if \textsc{QuIntSynt} returns no interpretation, then we can immediately exclude the searched region from further exploration and thus no new points are added to $W$ in this case. For example, as shown in \Cref{fig:iter3}, if \textsc{QuIntSynt} found no Pareto-optimal interpretations in $R_3^{c''',e'''}$, then this region is excluded from the search and \Cref{alg:allintsynt} continues with the next available point in $W$.  

Next we show some important properties of \Cref{alg:allintsynt}.

\begin{lemma}[Soundness]
	For an instance $\langle \interps, \mathcal S, \corrmeasure, \explmeasure\rangle$ of the Pareto-optimal interpretation synthesis problem, if $(E,(c,e)) \in \textsc{ExplorePOI}(\interps, \mathcal S, \corrmeasure, \explmeasure)$, then $(c,e) \in \underset{E'\in \interps}{\max}^{\preceq} (\corrmeasure(f_{\interp'},\mathcal S), \explmeasure(\interp'))$.
\label{lem:sound}
\end{lemma}

In the rest of this section, we assume that each of the explainability measures has finitely many discrete values, as they are defined as floating points up to a certain precision. Thus, we obtain that the range of $\explmeasure$ is finite, which allows us to obtain the following results. %
\begin{lemma}[Completeness]
	For an instance $\langle \interps, \mathcal S, \corrmeasure, \explmeasure\rangle$ of the Pareto-optimal interpretation synthesis problem, if $(c,e) \in \underset{E'\in \interps}{\max}^\preceq (\corrmeasure(f_{\interp'},\mathcal S), \explmeasure(\interp'))$, then there is an interpretation $\interp$ with measures $(c,e)$ such that $(\interp,(c,e)) \in \textsc{ExplorePOI}(\interps, \mathcal S, \corrmeasure, \explmeasure)$.
\label{lem:comp}
\end{lemma}

We summarize the correctness result next which follows immediately from \Cref{lem:sound,lem:comp}. 

\begin{theorem}[Correctness of \Cref{alg:allintsynt}]
	For a class of interpretations $\interps$, a finite set of samples $\mathcal S$, and measures $\corrmeasure$ and $\explmeasure$, the algorithm \textsc{ExplorePOI} terminates and returns a minimal representative set for $(\interps,\mathcal S, \corrmeasure, \explmeasure )$.
\end{theorem}

Algorithm {\textsc{ExplorePOI} solves the interpretation synthesis
  problem as a multi-objective optimization problem.  If we were to
  solve the same problem using single-objective optimization, it would
  be necessary to combine the accuracy and explainability measures for
  every interpretation to yield a single hybrid measure.  Let
  $\lambda: \mathbb{R} \times \mathbb{R} \rightarrow \mathbb{R}$ be a function
  that yields such a measure.  Since higher values of $c$ and $e$
  always increase the desirability of an interpretation, we require
  $\lambda$ to be {\em strictly increasing}, i.e., $(c, e) \prec (c', e') \implies
  \lambda(c, e) < \lambda(c', e')$. %
  For example, $\lambda(c, e) = w_1\cdot c + w_2 \cdot e$ is a strictly increasing function for every $w_1, w_2 > 0$. Then, for any $(c,e)$ pair that is maximal wrt such a function $\lambda$, our algorithm can find an interpretation with this measure pair. Formally,

\begin{theorem}[Universality]
  For every strictly increasing function $\lambda:\mathbb{R}\times \mathbb{R}\rightarrow \mathbb{R}$ and
every $\langle \interps, \mathcal S, \corrmeasure, \explmeasure\rangle$ %
if $\interp \in \arg\max \limits_{\interp'\in \interps} (\lambda(\corrmeasure(f_{\interp'},\mathcal S), \explmeasure(\interp')))$, then there exists an interpretation ${\interp^\star}\in \interps$ such that (i) $\corrmeasure(f_\interp,\mathcal S) = \corrmeasure(f_{\interp^\star}, \mathcal S)$, (ii) $\explmeasure(\interp) = \explmeasure(\interp^\star)$, and
(iii) $(\interp^\star, (\corrmeasure(f_{\interp^\star}, \mathcal S), \explmeasure(\interp^\star))) \in \textsc{ExplorePOI}(\interps, \mathcal S, \corrmeasure, \explmeasure)$.
\label{lem:wt}
\end{theorem}

We conclude the section with some remarks on \Cref{alg:allintsynt}. 
\begin{remark}
\Cref{alg:allintsynt} can also be applied interactively as a conversation between synthesizer and user. Given a Pareto-optimal interpretation, the user may guide the search to interpretations that are more explainable or to those with more accuracy, until the user has found an optimal interpretation.  
\end{remark}

\begin{remark}
  Note that there might be multiple interpretations with the same pair $(c,e)$. In this case, \Cref{alg:allintsynt} will add only one of them as a representative interpretation, since the others are indistinguishable wrt correctness and explainability.  
\end{remark}

	Finally, we can also search for Pareto-optimal solutions based on regions solely bounded on the correctness measure. We choose to use bounds on the explainability measure, because the sample sets tend to be large and will result in much larger encodings.

\section{Statistical Guarantees for Black-box Models}

\label{sec:refinement}

In \Cref{sec:po-synthesis}, the correctness of an interpretation
$\interp$, defined using a measure $\corrmeasure$, was determined with
respect to a set of samples $\mathcal S$ obtained from the black-box
model $\mathcal B$.  Our approach guarantees that $\interp$ is optimal
for $\mathcal S$ and the measure $\corrmeasure$.  Our ultimate goal
is, however, to synthesize an interpretation $\interp$ that is optimal
with respect to the entire black-box model $\mathcal B$, i.e.,
w.r.t. the set $\mathcal S_{\mathcal B} = \{(i,o) \mid
f_\blackbox(i)=o\}$. Obtaining an exhaustive set of samples from a
black-box model is often not practical. The question that we,
therefore, raise in this section is, how large a set of samples
$\mathcal S$ must be such that it is not \emph{misleading}, i.e.,
optimal interpretations synthesized by our approach for the set of
samples do not overfit the set, and thus the guarantees obtained over
$\mathcal S$ can be adopted for $\mathcal S_{\mathcal B}$.

The answer to the latter question lies in the theory of \emph{Probably
Approximately Correct Learnability} (PAC) \cite{understandingML}.
Specifically, a class of hypotheses (interpretations) $\interps$ over
inputs $\mathcal I$ and outputs $\mathcal O$ is PAC-learnable with
respect to the set $Z= \mathcal I \times \mathcal O$ and a loss
function $\ell \colon (\mathcal I \rightarrow \mathcal O) \times Z
\rightarrow [0,1]$, if there exists a function $m_\interps\colon
(0,1)^2 \rightarrow \mathbb N$ and a learning algorithm with the
following property: For every $\epsilon$, $\delta \in (0,1)$ and for
every distribution $D$ over $Z$, when running the learning algorithm
on $m\ge m_\interps(\epsilon, \delta)$ i.i.d. samples generated by
$D$, the algorithm returns a hypothesis $\interp$ such that, with
probability (confidence) of at least $1-\delta$, $L_D(f_\interp) -
\underset{\interp' \in~\interps}{\min}~L_D(f_{\interp'}) \leq
\epsilon$, where $L_D(f_E) = \mathbb E_{z\sim D} [\ell(f_E,z)]$.
Furthermore, an algorithm that chooses an interpretation $\interp \in
\interps$ that minimizes $\frac{\sum_{z \in \mathcal S}
  \ell(f_\interp, z)}{|\mathcal S|}$ suffices for the learning
algorithm in the above definition~\cite{understandingML}.

For our purposes, we assume that the correctness measure
$\Delta_{\mathcal C}$ has range $[0,1]$ (achievable by normalization),
and use $1 - \Delta_{\mathcal C}$ for the loss function $\ell$
referred to above.  Thus, if $z = (i, o)$ is a sample, then
$\ell(f_\interp, z)$ is given by $1 - \Delta_{\mathcal C}(f_\interp,
\{(i,o)\})$.

It is known that every finite class of interpretations is
PAC-learnable due to the uniform convergence property
\cite{understandingML}.  In fact, the sample complexity, i.e., the
function $m_\interps$, can be determined in terms of $|\interps|$,
$\delta$ and $\epsilon$.  Under the standard \emph{realizability
assumption}, i.e $\interps$ includes an interpretation $E$ such that
$f_E$ implements the semantic function $f_{\mathcal B}$ of the
black-box, $m_\interps$ is bounded above by
$\lceil{\frac{\log{(|\interps|/ \delta)}}{\epsilon}}\rceil$. This
bound increases to $\lceil{\frac{2\log{(2|\interps|/
      \delta)}}{\epsilon^2}}\rceil$ sans the realizability
assumption\cite{understandingML}.

Using the above bounds for the sample size results in interpretations
that are very close to the optimal interpretation within the class of
interpretations with high probability, yet does not necessarily mean
it is very close to the black-box model. The latter depends highly on
the class of interpretations.
Furthermore, despite the big advantage of obtaining optimality guarantee on the synthesized interpretation, and without
sampling the entire set $\mathcal S_\mathcal B$, the price for this
guarantee is that we may have to work with an increased size of the
sample set $\mathcal S$. In general, this affects the scalability of
our synthesis procedure, since size of the weighted MaxSAT formula
increases linearly with $|\mathcal S|$. This can limit how
small $\delta$ and $\epsilon$ can be in practice.  Nevertheless, as we
show in Section~\ref{sec:experiments}, we are able to use fairly
small values of $\delta$ and $\epsilon$ in our experiments.

\section{Evaluation}
\label{sec:experiments}

\paragraph{Benchmarks.} We apply our approach to three black-box models: a \emph{decision module} for predicting the performance of a perception module in an airplane (AP), a \emph{bank loan predictor} (BL), and a \emph{solvability predictor} (TP). 

 The decision module predicts, based on the time of day, the cloud types, and initial positioning of an airplane on a runway, whether a perception module used by the plane can be trusted to behave correctly. The decision module is an implementation of a decision tree that was trained on data collected from 200 simulations, using the XPlane (\url{x-plane.org}) simulator. 

The bank loan  predictor is a deep neural network that was trained on synthetic data that we created. The training set included 100000 entries chosen such that majority of people with age between 18 to 29 years, and those with age between 30 and 49 years but with income less than \$6000, were denied the loan. The network has five dense fully connected hidden layers with 200  ReLU's each, in addition to a Softmax layer and the output layer of two nodes.

The solvability predictor is a neural network built to predict the solvability of first-order formulas by a theorem prover with respect to percentage of unit clauses and average clause length in a formula. The network had three hidden dense fully connected layers each with 200 ReLU's.  The data used to train the neural network can be found on the UCI machine learning repository~\cite{theoremProver}.  We used the data for heuristic H1 from~\cite{theoremProver}, thus predicting solvability for H1.

\paragraph{Experiments and setup} We conducted two types of experiments: (1) Application of exploration algorithm on the three benchmarks (2) performance evaluation of \textsc{QuIntSynt}. 
 The MaxSAT engine used an implementation of RC2 in PySAT \cite{imms-sat18,DBLP:journals/jsat/IgnatievMM19}.  
All experiments were conducted on a 2.4GHz Quad-core machine with 8GB of RAM. More detailed experiments and results are in the Appendix.

\paragraph{Exploring the Pareto-optimal space}

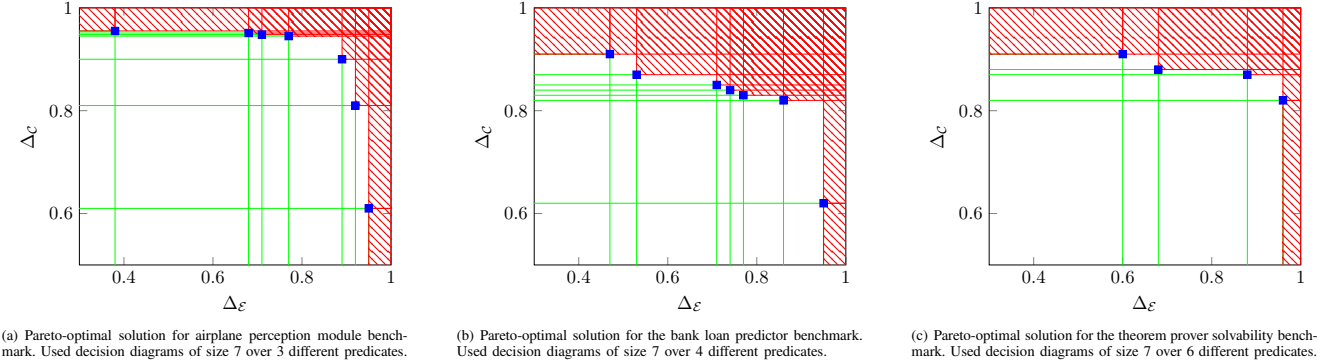
\begin{figure*}
\centering
\scalebox{0.7}[0.7]{
\subfigure[Pareto-optimal solution for airplane perception module benchmark. Used decision diagrams of size 7 over 3 different predicates.]{
\label{fig:paretoAP}
\begin{tikzpicture}
	\begin{axis}[xmin=0.30, xmax=1, ymin=0.5, ymax=1, samples=50,xlabel={$\explmeasure$},
    ylabel={$\corrmeasure$}]
    \addplot[blue, mark=square*] coordinates {(0.89,0.9)} node[below left] {\small};
    \draw[pattern=dots, pattern color=green, draw=green] (0.89,0.9) -- (0,0.9);
    \draw[pattern=dots, pattern color=green, draw=green] (0.89,0.9) -- (0.89,0);
    \draw[pattern=north west lines, pattern color=red, draw=red] (0.89,0.9) rectangle (1,1);
    
    \addplot[blue, mark=square*] coordinates {(0.92,0.81)} node[below left] {\small};
    \draw[pattern=dots, pattern color=green, draw=green] (0.92,0.81) -- (0,0.81);
    \draw[pattern=dots, pattern color=green, draw=green] (0.92,0.81) -- (0.92,0);
    \draw[pattern=north west lines, pattern color=red, draw=red] (0.92,0.81) rectangle (1,1);
    
    \addplot[blue, mark=square*] coordinates {(0.95,0.61)} node[below left] {\small};
    \draw[pattern=dots, pattern color=green, draw=green] (0.95,0.61) -- (0,0.61);
    \draw[pattern=dots, pattern color=green, draw=green] (0.95,0.61) -- (0.95,0);
    \draw[pattern=north west lines, pattern color=red, draw=red] (0.95,0.61) rectangle (1,1);
    \draw[pattern=north west lines, pattern color=red, draw=red] (0.95,0.61) rectangle (1,0);
    
    \addplot[blue, mark=square*] coordinates {(0.77,0.945)} node[below left] {\small};
    \draw[pattern=dots, pattern color=green, draw=green] (0.77,0.945) -- (0,0.945);
    \draw[pattern=dots, pattern color=green, draw=green] (0.77,0.945) -- (0.77,0);
    \draw[pattern=north west lines, pattern color=red, draw=red] (0.77,0.945) rectangle (1,1);
    
    \addplot[blue, mark=square*] coordinates {(0.71,0.948)} node[below left] {\small};
    \draw[pattern=dots, pattern color=green, draw=green] (0.71,0.948) -- (0,0.948);
    \draw[pattern=dots, pattern color=green, draw=green] (0.71,0.948) -- (0.71,0);
    \draw[pattern=north west lines, pattern color=red, draw=red] (0.71,0.948) rectangle (1,1);
    
    \addplot[blue, mark=square*] coordinates {(0.68,0.951)} node[below left] {\small};
    \draw[pattern=dots, pattern color=green, draw=green] (0.68,0.951) -- (0,0.951);
    \draw[pattern=dots, pattern color=green, draw=green] (0.68,0.951) -- (0.68,0);
    \draw[pattern=north west lines, pattern color=red, draw=red] (0.68,0.951) rectangle (1,1);
    
    \addplot[blue, mark=square*] coordinates {(0.38,0.955)} node[below left] {\small};
    \draw[pattern=dots, pattern color=green, draw=green] (0.38,0.955) -- (0,0.955);
    \draw[pattern=dots, pattern color=green, draw=green] (0.38,0.955) -- (0.38,0);
    \draw[pattern=north west lines, pattern color=red, draw=red] (0.38,0.955) rectangle (1,1);
    \draw[pattern=north west lines, pattern color=red, draw=red] (0.38,0.955) rectangle (0,1);

	\end{axis}
	
\end{tikzpicture}
}
}
\quad
\scalebox{0.7}[0.7]{
\subfigure[Pareto-optimal solution for the bank loan predictor benchmark. Used decision diagrams of size 7 over 4 different predicates.]{
\label{fig:paretoBL}
\begin{tikzpicture}
	\begin{axis}[xmin=0.30, xmax=1, ymin=0.5, ymax=1, samples=50,xlabel={$\explmeasure$},
    ylabel={$\corrmeasure$}]
    \addplot[blue, mark=square*] coordinates {(0.95,0.62)} node[below left] {\small};
    \draw[pattern=dots, pattern color=green, draw=green] (0.95,0.62) -- (0,0.62);
    \draw[pattern=dots, pattern color=green, draw=green] (0.95,0.62)-- (0.95,0);
    \draw[pattern=north west lines, pattern color=red, draw=red] (0.95,0.62) rectangle (1,1);
    \draw[pattern=north west lines, pattern color=red, draw=red] (0.95,0.62) rectangle (1,0);
    
    \addplot[blue, mark=square*] coordinates {(0.86,0.82)} node[below left] {\small};
    \draw[pattern=dots, pattern color=green, draw=green] (0.86,0.82) -- (0,0.82);
    \draw[pattern=dots, pattern color=green, draw=green] (0.86,0.82)-- (0.86,0);
    \draw[pattern=north west lines, pattern color=red, draw=red] (0.86,0.82) rectangle (1,1);
    
    \addplot[blue, mark=square*] coordinates {(0.77,0.83)} node[below left] {\small};
    \draw[pattern=dots, pattern color=green, draw=green] (0.77,0.83) -- (0,0.83);
    \draw[pattern=dots, pattern color=green, draw=green] (0.77,0.83)-- (0.77,0);
    \draw[pattern=north west lines, pattern color=red, draw=red] (0.77,0.83) rectangle (1,1);

	\addplot[blue, mark=square*] coordinates {(0.74,0.84)} node[below left] {\small};
    \draw[pattern=dots, pattern color=green, draw=green] (0.74,0.84) -- (0,0.84);
    \draw[pattern=dots, pattern color=green, draw=green] (0.74,0.84)-- (0.74,0);
    \draw[pattern=north west lines, pattern color=red, draw=red] (0.74,0.84) rectangle (1,1);
    
    \addplot[blue, mark=square*] coordinates {(0.71,0.85)} node[below left] {\small};
    \draw[pattern=dots, pattern color=green, draw=green] (0.71,0.85) -- (0,0.85);
    \draw[pattern=dots, pattern color=green, draw=green] (0.71,0.85)-- (0.71,0);
    \draw[pattern=north west lines, pattern color=red, draw=red] (0.71,0.85) rectangle (1,1);
    
    \addplot[blue, mark=square*] coordinates {(0.53,0.87)} node[below left] {\small};
    \draw[pattern=dots, pattern color=green, draw=green] (0.53,0.87) -- (0,0.87);
    \draw[pattern=dots, pattern color=green, draw=green] (0.53,0.87)-- (0.53,0);
    \draw[pattern=north west lines, pattern color=red, draw=red] (0.53,0.87) rectangle (1,1);
    
    \addplot[blue, mark=square*] coordinates {(0.47,0.91)} node[below left] {\small};
    \draw[pattern=dots, pattern color=green, draw=green] (0.47,0.91) -- (0,0.91);
    \draw[pattern=dots, pattern color=green, draw=green] (0.47,0.91)-- (0.47,0);
    \draw[pattern=north west lines, pattern color=red, draw=red] (0.47,0.91) rectangle (1,1);
	\draw[pattern=north west lines, pattern color=red, draw=red] (0.47,0.91) rectangle (0,1);

    \end{axis}
\end{tikzpicture}
}
}
\quad
\scalebox{0.7}[0.7]{
\subfigure[Pareto-optimal solution for the theorem prover solvability benchmark. Used decision diagrams of size 7 over 6 different predicates.]{
\label{fig:paretoTP}
	\begin{tikzpicture}
		\begin{axis}[xmin=0.30, xmax=1, ymin=0.5, ymax=1, samples=50,xlabel={$\explmeasure$},
    ylabel={$\corrmeasure$}]
    \addplot[blue, mark=square*] coordinates {(0.6,0.91) } node[below left] {\small};
    \draw[pattern=dots, pattern color=green, draw=green] (0.6,0.91) -- (0,0.91);
    \draw[pattern=dots, pattern color=green, draw=green] (0.6,0.91) -- (0.6,0);
    \draw[pattern=north west lines, pattern color=red, draw=red] (0.6,0.91) rectangle (1,1);
    \draw[pattern=north west lines, pattern color=red, draw=red] (0.6,0.91)  rectangle (0,1);

	\addplot[blue, mark=square*] coordinates {(0.68,0.88) } node[below left] {\small};
    \draw[pattern=dots, pattern color=green, draw=green]  (0.68,0.88) -- (0,0.88);
    \draw[pattern=dots, pattern color=green, draw=green] (0.68,0.88) -- (0.68,0);
    \draw[pattern=north west lines, pattern color=red, draw=red] (0.68,0.88)  rectangle (1,1);
    
    \addplot[blue, mark=square*] coordinates {(0.88,0.87) } node[below left] {\small};
    \draw[pattern=dots, pattern color=green, draw=green] (0.88,0.87) -- (0,0.87);
    \draw[pattern=dots, pattern color=green, draw=green] (0.88,0.87) -- (0.88,0);
    \draw[pattern=north west lines, pattern color=red, draw=red] (0.88,0.87) rectangle (1,1);

    \addplot[blue, mark=square*] coordinates {(0.96,0.82) } node[below left] {\small};
    \draw[pattern=dots, pattern color=green, draw=green] (0.96,0.82) -- (0,0.82);
    \draw[pattern=dots, pattern color=green, draw=green] (0.96,0.82) -- (0.96,0);
    \draw[pattern=north west lines, pattern color=red, draw=red] (0.96,0.82) rectangle (1,1);
    \draw[pattern=north west lines, pattern color=red, draw=red] (0.96,0.82)  rectangle (1,0);

    \end{axis}
	\end{tikzpicture}
}
}

\caption{Exploring Pareto-optimal solutions for three benchmarks. The size of the sample sets used for constructing interpretations was computed based on confidence values $\delta=0.05$ and error margin $\epsilon=0.05$, as well as the size of the class of interpretation in each benchmark.}
\label{fig:curves}
\end{figure*}

We ran our approach on the three benchmarks mentioned above. Here we
made the realizability assumption referred to in
Section~\ref{sec:refinement}, and used confidence measure
$\delta=0.05$ and error margin $\epsilon=0.05$ to determine the size
of sample set to use for each benchmark (size of sample sets are given
in
\Cref{tab:perf:qsynth}). \Cref{fig:paretoAP,fig:paretoBL,fig:paretoTP}
show the measures of the Pareto-optimal interpretations found by our
exploration algorithm. We used accuracy for correctness, and
explainability measure that favored decision diagrams of smaller size
and predicates with a fewer number of branchings.

For all three benchmarks we found a variety of Interpretations with tradeoffs on the correctness and explainability measure, reflected by the outermost point in each plot. The exploration algorithm shows, that searching only for interpretation that are solely optimal in size or in accuracy may result in unfavorable solutions. For example,  in \Cref{fig:paretoAP} we see that the interpretation with highest accuracy has very low explainability. However, a very small tradeoff in accuracy resulted in much more explainable interpretations. 

\paragraph{Performance}
Table \ref{tab:perf:qsynth} presents our results on each benchmark and gives the confidence value $\delta$, error rate $\epsilon$ and the number of samples $|S|$ used for each run; the number of Pareto-optimal points (PO), total number of points (TNP) and minimum, maximum and median times obtained. The number shown in parenthesis next to each benchmark is the number of predicates used. 
From Table \ref{tab:perf:qsynth} we can see that the number of Pareto-optimal (PO) points is considerably smaller than the total number of points (TNP). 
The minimum time taken to find an interpretation was less than $3$ seconds for all benchmarks, but there were a few points in the Pareto-optimal space where finding an interpretation took considerably more time - given by the maximum times. For most Pareto-optimal points though, the time taken to the find an interpretation was less than 20 seconds, as demonstrated by the median values. If an interpretation did not exist for a combination of correctness and explanability measures, the {\textsc MaxSAT} solver returned UNSAT in less than a second in all performance runs.

\begin{table}[t]
\caption{Performance of QuIntSynt: Exploration of the entire Pareto-optimal Space}
\label{tab:perf:qsynth}	
\centering
\resizebox{\columnwidth}{!}{
\begin{tabular}{|c|c|c|c|c|c|c|c| }
\hline
	{Bench} &  &  & {Explored} & {min} & {max} & {median} & unsat  \\
	{mark} & {$\delta$,$\epsilon$} & {$|\mathcal S|$} &   (PO, TNP) & time (s) & time (s) &  time (s) &time (s) \\
\hline
	{Theorem} &{0.05, 0.05} & 338 & 4, 20 & 0.767 & 3.392 & 1.138 & $<1$\\
\cline{2-8} 
	  {Prover (6)} & {0.05, 0.03} & 703 & 3, 28 & 2.051 & 18.148 & 3.643 &$<1$ \\
\hline
	{Air} &  {0.05, 0.05} & 333 & 7, 25 & 1.709 & 388.527 & 5.696 &  $<1$ \\
\cline{2-8} 
	  {plane (3)} & {0.05, 0.03} & 555 & 5, 26 & 2.513 & 616.520 & 11.222 &  $<1$  \\
\hline
	{Bank} & {0.05, 0.05} & 365 & 7, 27 & 1.927 & 387.599 & 8.975  & $<1$ \\
\cline{2-8} 
	{Loan (4)} &  {0.05, 0.03} & 608 & 4, 27 & 2.855 & 1299.196  & 17.998  & $<1$ \\
\hline
\end{tabular}
}
\end{table}

As none of the other interpretation synthesis tools in the literature compute the set of all Pareto optimal interpretations, we omit comparison with other tools (any such comparison wouldn't be fair, especially when using different notions for explainability). However, to understand if the variation in running times is inherent to the problem, we performed a similar experiment with MinDS, a tool for learning decision sets~\cite{opt-dec-sets}.  In MinDS, correctness and explainability are combined in a single objective and the contribution of the explainability measure is governed by a parameter $\lambda$. We ran MinDS for $15$ values of $\lambda$ and found interpretations for all these values. We observed again (Table \ref{tab:perf:minds}) that the time taken to find interpretations for some $\lambda$ was much more than others. 

Note that unlike in our approach, running MinDS in this manner does not guarantee that the entire Pareto-optimal space of interpretations has been obtained. Finding all Pareto optimal points by varying the weights of explainability and correctness measures is also not feasible, since this requires trying out all (infinitely many) weight combinations. While some of decision sets learned by MinDS were indeed semantically equivalent to some of the Pareto-optimal interpretations synthesized by our approach, some interpretations that our methods found did not have a decision set counterpart within  the range of weights we experimented on. We especially, emphasize that running approaches like MinDS or in general approaches based on a single objective function may even result in the same interpretation for different weights. This can be avoided using our exploration method. 

\begin{table}[h]
\caption{Illustrating variation in running times even on non-exhaustive Pareto search with MinDS}
\label{tab:perf:minds}	
\centering
	\scalebox{1}[1]{
\begin{tabular}{|c|c|c|c|c|c| }
\hline
	{Bench} &  &  & {min} & {max} & {median}  \\
	{mark} & {$\delta$,$\epsilon$} & {$|\mathcal S|$} &    time (s) & time (s) &  time (s)  \\
\hline
	{Theorem} &{0.05, 0.05} & 338 &  0.707 & 0.813 & 0.719\\

\cline{2-6} 
	  {Prover (6)} & {0.05, 0.03} & 703 & 0.687 & 0.798 & 0.725 \\
\hline
	{Air} &  {0.05, 0.05} & 333 & 0.771 & 364.456 & 7.603 \\

\cline{2-6} 
	  {plane (3)} & {0.05, 0.03} & 555 & 0.748 & 757.639 & 9.687\\

\hline
	{Bank} & {0.05, 0.05} & 365 &  0.744 & 25.819  & 1.165  \\
\cline{2-6} 
	{Loan (4)} &  {0.05, 0.03} & 608 & 0.738 & 52.388 & 0.841  \\
\hline
\end{tabular}
}
\end{table}

\section{Related Work}

There is a large body of work on interpreting black-box models, where a dominant paradigm is to generate labeled data samples and obtain an interpretable model representation in terms of input features, some of which were discussed in the introduction. In some applications, the aim is to explain the output of a black-box model in the neighbourhood of a specific input, and specialized techniques~\cite{lime,lore,anchors, shap,armando} give such local and robust explanations. Other applications use techniques like model distillation (in the form of decision trees~\cite{palm,treepan,dectext,krishnan,evolve-dt}), counterfactual explanations~\cite{molnar2019}.  For further information on these techniques, we refer to reader to the excellent surveys in~\cite{survey-paper,adadi2018peeking}.

The work in \cite{opt-dec-sets, hu19} comes closest to ours. In \cite{opt-dec-sets}, the authors encode the problem of finding an interpretation as optimal decision sets (to a weighted \textsc{MaxSat} formulation). They present two variants: optimize on accuracy (100\%) while constraining the explanability (number of literals) and directly minimize the size of decision sets at the cost of accuracy. In \cite{hu19}, sparse optimal decision trees are built using an objective function which combines misclassification rate and number of leaves. Solutions to these give a single point of the optimized function in the Pareto-optimal space and hence a single value for the correctness and explainability measures. %

Our Pareto-optimal interpretation synthesis problem formulation (Definition~\ref{def:qsygus}) can also be related to Structural Risk Minimization (SRM), which has been well-studied in the literature. Like in SRM, we have two orthogonal measures -- one that depends only on the structure/complexity of the hypothesis/interpretation, and the other that depends on how well the hypothesis/interpretation ``explains'' the given sample set. The SRM formulation (e.g., as defined in~\cite{understandingML}, Section~7.2) effectively combines these two measures into one and treats the problem as a single-objective optimization problem. %
In contrast, our Pareto-optimal synthesis problem is inherently a multi-objective optimization problem. As mentioned in the introduction, such a multi-objective optimization problem cannot be reduced to a single-objective optimization problem in general, without potentially excluding some (possibly important) solutions. Furthermore, we wish to compute minimal representative sets of Pareto-optimal interpretations (Definition~\ref{def:minrepset}). Since some Pareto-optimal solutions can get excluded in going from multi to single-objective optimization, the minimal representative set (or argmin) computed by the SRM approach can indeed differ from the set of solutions for our formulation.

\section{Conclusion and Futurework}
We have presented a new approach to automatically generate a complete set of Pareto-optimal interpretations for black-box ML models, which works in the absence of training or test data sets. Our interpretations, as decision diagrams, satisfy optimality conditions and provide formal guarantees on the tradeoff between accuracy and explainability.
We present an empirical evaluation demonstrating that our approach produces compact, accurate explanatory interpretations for neural networks used for applications such as autonomous plane taxiing, predicting bank loans, classifying theorem-provers and shows the value of the multi-objective approach.%

Our main contributions lie in investigating algorithmic approaches to solve the mentioned problems when the space of possible interpretations is finite. However, we note that finiteness of the hypothesis class doesn't immediately yield a practical algorithm for solving the problem. Indeed, the hypothesis class can be finite yet combinatorially large, as is the case in our examples. A naive enumeration-based algorithm is infeasible in practice in such cases. The weighted MaxSAT encoding allows us to solve this problem symbolically by leveraging significant recent advances in MaxSAT solving that scale to very large solution spaces. Using a finite, yet large hypothesis class permits us to strike a balance between generality and practical efficiency of our approach. Our overall encoding strategy, i.e. partitioning the encoding into four parts and using weights for specific variables in the encoding, is applicable in other settings like optimization modulo theories (OMT) that go beyond weighted MaxSAT, if such encodings are necessary for the underlying class of interpretations and measures.

An interesting avenue for futurework would be to see if this approach can be extended to work with interpretation classes of infinite cardinality but finite VC dimension. While the overall problem formulation, the notions of Pareto-optimality of explanations, and our algorithm for finding representative sets of explanations easily adapt to the setting of infinite classes of interpretations, it would possibly require going beyond weighted MaxSAT to find a Pareto-optimal explanation in a given interval of explainability scores. Using an encoding in optimization modulo theories (OMT) is a promising direction for such a generalization.

\vskip 0.5cm
\noindent {\bf Acknowledgments.} This work is partially supported by NSF grants 1545126 (VeHICaL), 1646208 and 1837132, by the DARPA contracts FA8750-18-C-0101 (AA) and FA8750-20-C-0156 (SDCPS), by Berkeley Deep Drive, and by Toyota under the iCyPhy center. We would also like to express our gratitude to the anonymous reviewers for their in-depth reviews, constructive suggestions and various pointers. 

\bibliography{ref.bib}

\begin{thebibliography}{10}

\bibitem{GDPR}
{General Data Protection Regulation (GDPR)}.
\newblock \url{https://gdpr.eu/}, 2018.

\bibitem{adadi2018peeking}
Amina Adadi and Mohammed Berrada.
\newblock {Peeking inside the black-box: A survey on Explainable Artificial
  Intelligence (XAI)}.
\newblock {\em IEEE Access}, 6:52138--52160, 2018.

\bibitem{learnopt-aai20}
Ga{\"{e}}l Aglin, Siegfried Nijssen, and Pierre Schaus.
\newblock Learning {O}ptimal {D}ecision {T}rees {U}sing {C}aching
  {B}ranch-and-{B}ound {S}earch.
\newblock In {\em {AAAI} 2020}, pages 3146--3153. {AAAI} Press, 2020.

\bibitem{alipanahi2015predicting}
Babak Alipanahi, Andrew Delong, Matthew~T Weirauch, and Brendan~J Frey.
\newblock {Predicting the sequence specificities of {DNA}-and {RNA}-binding
  proteins by deep learning}.
\newblock {\em Nature biotechnology}, 2015.

\bibitem{optimaldt-aaai20}
Florent Avellaneda.
\newblock Efficient {I}nference of {O}ptimal {D}ecision {T}rees.
\newblock In {\em {AAAI} 2020}, pages 3195--3202. {AAAI} Press, 2020.

\bibitem{dectext}
Olcay Boz.
\newblock {Extracting Decision Trees from Trained Neural Networks}.
\newblock In {\em Proceedings of the Eighth ACM SIGKDD International Conference
  on Knowledge Discovery and Data Mining}, KDD ’02, New York, NY, USA, 2002.
  Association for Computing Machinery.

\bibitem{theoremProver}
James~P. Bridge, Sean~B. Holden, and Lawrence~C. Paulson.
\newblock {Machine Learning for First-Order Theorem Proving - Learning to
  Select a Good Heuristic}.
\newblock {\em J. Autom. Reasoning}, 53(2):141--172, 2014.
\newblock
  \url{https://archive.ics.uci.edu/ml/datasets/First-order+theorem+proving}.

\bibitem{treepan}
Mark~W. Craven and Jude~W. Shavlik.
\newblock {Extracting Tree-Structured Representations of Trained Networks}.
\newblock In {\em Proceedings of the 8th International Conference on Neural
  Information Processing Systems}, NIPS’95, page 24–30, Cambridge, MA, USA,
  1995. MIT Press.

\bibitem{dahl2013large}
George~E Dahl, Jack~W Stokes, Li~Deng, and Dong Yu.
\newblock {Large-scale malware classification using random projections and
  neural networks}.
\newblock In {\em Proceedings of the IEEE International Conference on
  Acoustics, Speech and Signal Processing (ICASSP)}, pages 3422--3426. IEEE,
  2013.

\bibitem{fremont-cav20}
Daniel~J. Fremont, Johnathan Chiu, Dragos~D. Margineantu, Denis Osipychev, and
  Sanjit~A. Seshia.
\newblock Formal analysis and redesign of a neural network-based aircraft
  taxiing system with {VerifAI}.
\newblock In {\em 32nd International Conference on Computer Aided Verification
  (CAV)}, July 2020.

\bibitem{lore}
Riccardo Guidotti, Anna Monreale, Salvatore Ruggieri, Dino Pedreschi, Franco
  Turini, and Fosca Giannotti.
\newblock Local {R}ule-{B}ased {E}xplanations of {B}lack {B}ox {D}ecision
  {S}ystems.
\newblock {\em CoRR}, abs/1805.10820, 2018.

\bibitem{survey-paper}
Riccardo Guidotti, Anna Monreale, Salvatore Ruggieri, Franco Turini, Fosca
  Giannotti, and Dino Pedreschi.
\newblock {A Survey of Methods for Explaining Black Box Models}.
\newblock {\em ACM Comput. Surv.}, 51(5), August 2018.

\bibitem{hu19}
Xiyang Hu, Cynthia Rudin, and Margo Seltzer.
\newblock Optimal {S}parse {D}ecision {T}rees.
\newblock In {\em Advances in Neural Information Processing Systems (NeurIPS)},
  2019.

\bibitem{imms-sat18}
Alexey Ignatiev, Antonio Morgado, and Joao Marques{-}Silva.
\newblock {PySAT:} {A} {Python} toolkit for prototyping with {SAT} oracles.
\newblock In {\em SAT}, pages 428--437, 2018.

\bibitem{DBLP:journals/jsat/IgnatievMM19}
Alexey Ignatiev, Ant{\'{o}}nio Morgado, and Jo{\~{a}}o Marques{-}Silva.
\newblock {RC2:} an efficient {MaxSAT} solver.
\newblock {\em J. Satisf. Boolean Model. Comput.}, 11(1):53--64, 2019.

\bibitem{janota20}
Mikol{\'{a}}s Janota and Ant{\'{o}}nio Morgado.
\newblock {SAT}-{B}ased {E}ncodings for {O}ptimal {D}ecision {T}rees with
  {E}xplicit {P}aths.
\newblock In Luca Pulina and Martina Seidl, editors, {\em Theory and
  Applications of Satisfiability Testing - {SAT} 2020}, volume 12178 of {\em
  Lecture Notes in Computer Science}, pages 501--518. Springer, 2020.

\bibitem{jha-jar19}
Susmit Jha, Tuhin Sahai, Vasumathi Raman, Alessandro Pinto, and Michael
  Francis.
\newblock {Explaining AI Decisions Using Efficient Methods for Learning Sparse
  Boolean Formulae}.
\newblock {\em J. Autom. Reasoning}, 63(4):1055--1075, 2019.

\bibitem{evolve-dt}
U.~{Johansson} and L.~{Niklasson}.
\newblock Evolving decision trees using oracle guides.
\newblock In {\em 2009 IEEE Symposium on Computational Intelligence and Data
  Mining}, pages 238--244, 2009.

\bibitem{PayPal}
Eric Knorr.
\newblock {How PayPal beats the bad guys with machine learning}.
\newblock
  http://www.infoworld.com/article/2907877/machine-learning/how-paypal-reduces-fraud-with-machine-learning.html,
  2015.

\bibitem{krishnan}
R.~Krishnan, G.~Sivakumar, and P.~Bhattacharya.
\newblock Extracting decision trees from trained neural networks.
\newblock {\em Pattern Recognition}, 32(12):1999 -- 2009, 1999.

\bibitem{palm}
Sanjay Krishnan and Eugene Wu.
\newblock {PALM}: Machine learning explanations for iterative debugging.
\newblock In {\em Proceedings of the 2nd Workshop on Human-In-the-Loop Data
  Analytics}, HILDA’17, New York, NY, USA, 2017. Association for Computing
  Machinery.

\bibitem{shap}
Scott~M Lundberg and Su-In Lee.
\newblock A {U}nified {A}pproach to {I}nterpreting {M}odel {P}redictions.
\newblock In I.~Guyon, U.~V. Luxburg, S.~Bengio, H.~Wallach, R.~Fergus,
  S.~Vishwanathan, and R.~Garnett, editors, {\em Advances in Neural Information
  Processing Systems 30}, pages 4765--4774. Curran Associates, Inc., 2017.

\bibitem{forbes2019mortgage}
Douglas Merrill.
\newblock {AI} is coming to take your mortgage woes away.
\newblock
  \url{https://www.forbes.com/sites/douglasmerrill/2019/04/04/ai-is-coming-to-take-your-mortgage-woes-away/},
  April 2019.

\bibitem{molnar2019}
Christoph Molnar.
\newblock {\em Interpretable {M}achine {L}earning}.
\newblock 2019.
\newblock \url{https://christophm.github.io/interpretable-ml-book/}.

\bibitem{nina18}
Nina Narodytska, Alexey Ignatiev, Filipe Pereira, and Jo{\~{a}}o
  Marques{-}Silva.
\newblock Learning {O}ptimal {D}ecision {T}rees with {SAT}.
\newblock In J{\'{e}}r{\^{o}}me Lang, editor, {\em International Joint
  Conference on Artificial Intelligence, {IJCAI} 2018}. ijcai.org, 2018.

\bibitem{NVIDIATegra}
NVIDIA.
\newblock Nvidia tegra drive px: Self-driving car computer, 2015.

\bibitem{lime}
Marco~Tulio Ribeiro, Sameer Singh, and Carlos Guestrin.
\newblock “{W}hy {S}hould {I} {T}rust {Y}ou?”: Explaining the {P}redictions
  of {A}ny {C}lassifier.
\newblock In {\em Knowledge Discovery and Data Mining}, KDD ’16. Association
  for Computing Machinery, 2016.

\bibitem{anchors}
Marco~Tulio Ribeiro, Sameer Singh, and Carlos Guestrin.
\newblock {Anchors: High-Precision Model-Agnostic Explanations}.
\newblock In {\em AAAI Conference on Artificial Intelligence}, 2018.

\bibitem{understandingML}
Shai Shalev-Shwartz and Shai Ben-David.
\newblock {\em {Understanding Machine Learning: From Theory to Algorithms}}.
\newblock Cambridge University Press, USA, 2014.

\bibitem{sirignano2016deep}
Justin Sirignano, Apaar Sadhwani, and Kay Giesecke.
\newblock Deep learning for mortgage risk, 2016.

\bibitem{introToDataMining}
Pang{-}Ning Tan, Michael~S. Steinbach, and Vipin Kumar.
\newblock {\em {Introduction to Data Mining}}.
\newblock Addison-Wesley, 2005.

\bibitem{verhaeghe20}
H{\'{e}}l{\`{e}}ne Verhaeghe, Siegfried Nijssen, Gilles Pesant, Claude{-}Guy
  Quimper, and Pierre Schaus.
\newblock Learning {O}ptimal {D}ecision {T}rees using {C}onstraint
  {P}rogramming (extended abstract).
\newblock In Christian Bessiere, editor, {\em Proceedings of the Twenty-Ninth
  International Joint Conference on Artificial Intelligence, {IJCAI} 2020},
  pages 4765--4769. ijcai.org, 2020.

\bibitem{dtip}
Sicco Verwer and Yingqian Zhang.
\newblock Learning {D}ecision {T}rees with {F}lexible {C}onstraints and
  {O}bjectives {U}sing {I}nteger {O}ptimization.
\newblock In Domenico Salvagnin and Michele Lombardi, editors, {\em Integration
  of AI and OR Techniques in Constraint Programming}, pages 94--103, Cham,
  2017. Springer International Publishing.

\bibitem{verwer19}
Sicco Verwer and Yingqian Zhang.
\newblock Learning {O}ptimal {C}lassification {T}rees {U}sing a {B}inary
  {L}inear {P}rogram {F}ormulation.
\newblock In {\em {AAAI} 2019}, pages 1625--1632. {AAAI} Press, 2019.

\bibitem{opt-dec-sets}
Jinqiang Yu, Alexey Ignatiev, Peter~J. Stuckey, and Pierre Le~Bodic.
\newblock {C}omputing {O}ptimal {D}ecision {S}ets with { SAT}.
\newblock In {\em Principles and Practice of Constraint Programming}, pages
  952--970, Cham, 2020. Springer International Publishing.

\bibitem{armando}
Xin Zhang, Armando Solar-Lezama, and Rishabh Singh.
\newblock {Interpreting Neural Network Judgments via Minimal, Stable, and
  Symbolic Corrections}.
\newblock In S.~Bengio, H.~Wallach, H.~Larochelle, K.~Grauman, N.~Cesa-Bianchi,
  and R.~Garnett, editors, {\em Advances in Neural Information Processing
  Systems 31}, pages 4874--4885. Curran Associates, Inc., 2018.

\end{thebibliography}
\bibliographystyle{plain}

%\newpage

\appendix

We provide additional details on our approach:
\begin{itemize}
	\item we include a detailed description of our weighted \textsc{MaxSat} encoding of the Pareto-optimal interpretation synthesis problem.
        \item we include the proof sketches for lemmas and theorems stated earlier.
	\item we provide more details about our benchmarks. 		
	\item we show some plots of execution times of explorations for the three models.
\end{itemize}

\subsection{\textsc{MaxSat} Encoding}
\label{app:encoding}
In the main paper, we gave a high-level description of our approach of solving the problem of synthesizing decision diagrams by encoding it as weighted maximum satisfiability (weighted \textsc{MaxSAT}). In this section, we discuss the encoding in detail.

For an instance $\langle \interps, \mathcal S, \corrmeasure, \explmeasure \rangle$ of the Pareto-optimal interpretation synthesis problem, the encoding is defined as a conjunction of formulae 
  $$\phi_{\langle \interps, \mathcal S, \corrmeasure, \explmeasure \rangle} = \phi_{\interps} \wedge \phi_{\mathcal S} \wedge \phi_{\corrmeasure} \wedge \phi_{\explmeasure}$$
   where

\paragraph{Encoding the space of intrpretations ($\phi_{\interps}$)} 

We choose \emph{bounded
multi-valued acyclic decision diagrams} with a specified maximum
number of decision nodes (Kindly note that other interpretations like decision trees, etc can be similarly encoded.). Let $\mathcal B$ be a black box defined over inputs $\mathcal I=\mathbb R^{\numofinputs}$ and outputs  $\mathcal O =\mathbb R^{\numofoutputs}$, for $\numofinputs,\numofoutputs \in \mathbb N$. 
The bounded decision diagram is an acyclic DAG consisting of $k$ internal nodes. 
The decision diagrams are further restricted by a finite set of decision predicates.  These predicates are defined as follows.  
Let 
 $ F = \{f_1, \dots,f_{\numoffeatures} \colon \mathcal I \rightarrow \mathbb R\}$ be a set of \emph{features}, which are functions that map input values of the black-box model to values in $\mathbb R$. 
Let further $B_{f_i}$ for $1\leq i\leq \numoffeatures $ be the sets $B_{f_i}=\{b_{f_i}^1,\dots, b_{f_{i}}^k \colon \mathbb R \rightarrow \{0,\dots,c_{f_i}^j\} \mid j \leq k, c_{f_i}^j \in \mathbb N\}$ that define partitions of the co-domain of each feature function $f_i$, which corresponds to the number of branches at a node in the decision diagram. We call an output of a partition a \emph{branching}. 
A decision predicate is then one of the functions $b_{f_i}^j \circ f_i$ that given input value $\vec{\iota} \in \mathcal I$ returns the branching $b_{f_i}^j( f_i(\vec{\iota}))$. 
We denote the set of predicates by $P= \{b \circ f \mid f\in F, b\in B_f \}$. Given a predicate $p \in P$, we define $F(p) = f$ and $B(p)=b$ as the feature and partition it is composed of, i.e., $b\circ f= p$.
Lastly let $L = \{\ell_1,\dots, \ell_{\numoflabels}\}$ be a set of output labels and $\sigma: \mathcal O \rightarrow L$ a function mapping outputs to labels. 
The formula $\phi_{\interps}$ encodes the space of decision diagrams  of $k$-nodes over a set of predicates $P$ and labels $L$ as the conjunction of the following constraints: 
\begin{itemize}
	\item Each internal node in the template is assigned exactly one predicate. 
    The encoding is given as:
$$\bigwedge \limits_{1\leq i\leq k} ~\bigvee \limits_{p \in  P} (\lambda_{i,p} \wedge \bigwedge \limits_{p'\not= p \in  P} \overline \lambda_{i,p'})$$ 
where $\lambda_{i,p}$ is true if node $i$ is assigned feature $p$ and false otherwise.
	\item From each internal node $i$, if predicate $p$ is assigned to $i$ then for every branching $c \in \textit{co-domain}(B(p))$,  there is an outgoing  transition from node $i$.
    This transition can go to another internal node or to a leaf representing a (label, bucket) pair. Transitions to internal nodes are unique - multiple transitions from a (feature, bucket) pair are disallowed to internal nodes.
    However, multiple transitions are possible for leaf nodes, specifically, to exactly one (label, bucket) pair for each label (For an input the decision diagram computes a value for each of the labels). The encoding below captures this requirement.
    The variable $\tau_{i,c,j}$ represents a transition from node $i$ to node $j$ labelled by bucket $b$.
  
\begin{equation*}
\begin{split}
	\bigwedge \limits_{1\leq i \leq k}~ \bigwedge 
	\limits_{p \in P}~ \lambda_{i,p} \rightarrow \\ \bigwedge \limits_{0\leq c\leq \max(B(p))} \big ( ~\bigvee 
	\limits_{\substack{j\in \{i+1,\dots,k\}\\ \cup L} } (\tau_{i,c,j}\wedge \bigwedge \limits_{ j' \not= j} \overline \tau_{i,c,j'}) \big)
\end{split}
\end{equation*}
Since the template is acylic, the nodes in the template are topologically ordered and a transition can only  go from node $i$ to either node $j$, ($j > i$), or  to a leaf representing a label.

	\item We need the following constraint for consistency: only branchings for the node predicate are allowed.  
    Let $c_\text{max}$ be the largest branching over all predicates, then we require that
	
$$\bigwedge \limits_{1\leq i\leq k}~ \bigwedge \limits_{p\in P} \lambda_{i,p} \rightarrow \bigwedge \limits_{\max(B(p)) < c \leq c_\text{max}} \bigwedge \limits_{\substack{j\in \{i+1,\dots,k\} \cup\\  L}}   \overline\tau_{i,c,j} $$

\end{itemize}

\paragraph{Encoding the relation between the template and the samples ($\phi_\mathcal S$)} The formula $\phi_\mathcal S$ establishes the relation between the samples and the template. 
 It encodes the definition of a matching variable $m_{i,s}$ at each node $i$
for each sample $s$ used for synthesis. 
The formula $\phi_\mathcal S$ sets these matching variables to true iff the interpretation (decision diagram) given by a satisfying assignment of $\phi_{\interps}$ produces a label of the output produced by the black box for the inputs in $s$.  
The encoding is done as follows. We encode a valid path from the leaves (labels) to the initial (root) node.
 We associate a variable $m_{i,s}$ with each node $i$ and sample $s$. This variable $m_{\ell, s}$  is true at a leaf, $\ell$, if a sample $s$ maps to the label which the leaf represents. 
 For any feature node $i$,  it is true at node $i$ if there is a valid path from $i$ to a leaf in the decision diagram.  %
 Let $func (s,  p/\ell, c)$ evaluate a sample on a predicate $p$ (label $\ell$, resp.) and return true if it falls in branching $c$ on the inputs in~$s$ (output of $s$ has label $\ell$. We omit c for labels). 
 \begin{itemize}
 	\item Encoding the leaves:  We have a $m_{\ell,s}$ variable for each sample $s$ and leaf $\ell$:

 $$\bigwedge \limits_{s \in {\mathcal S}} \bigwedge \limits_{\ell \in \mathcal L}  func(s, \ell) \leftrightarrow m_{\ell, s}$$

 	\item Next,
 $m_{i,s}$ at a template node $i$ is true if there is a transition from $i$ to $j$ on predicate $f$ and branching $c$ and the inputs of the sample $s$ match the ($p$,$b$) pair at $i$ and the path from $j$ is already valid, i.e., $m_{j,s}$ is true.

\begin{equation*}
\begin{split}
 		\bigwedge \limits_{s \in {\mathcal S}} \bigwedge \limits_{1\leq i\leq k} m_{i,s} \leftrightarrow \big( 
 			\bigvee \limits_{p\in P} \bigvee \limits_{c\leq \max(B(p))} ~
 			\lambda_{i,f}  \wedge  func(s, p, c)  \wedge \\
 			\bigwedge \limits_{\substack{j \in \{i+1,\dots,k\} \cup\\  L}}
            ( \tau_{i,b,j} \rightarrow m_{j,s)})
 			~\big)
\end{split}
\end{equation*}
\end{itemize}

In our weighted \textsc{MaxSAT} formulation, we require all clauses resulting from a Tseitin encoding of $\phi_{\interps} \wedge \phi_{\mathcal S}$ to be hard clauses. 
To complete the encoding we need the additional constraint for $\corrmeasure$ and $\explmeasure$, which will define the soft clauses we want to maximize the weights over.

\paragraph{Encoding the correctness measure ($\phi_{\corrmeasure}$)}
We require 
that $m_{1,s}$, for each sample $s$, to be true, i.e., 
$$\bigwedge \limits_{s\in \mathcal S} m_{1,s}$$

  This
additional constraint is added in the form of a  unit soft clause $m_{1,s}$ for each
sample $s$ with  weight set to $1$.  
Kindly note that $\phi_{\interps} \wedge \phi_{\mathcal S} $ is always satisfiable if we don't insist that at least one $m_{1,s}$ variable must be assigned $1$.  The variables $m_{1,s}$ correspond to the variables $m_{(i,o)}$ described in the paper. For other type of quantitative function, the user just needs to change the weights as we describe in the paper. 

\paragraph{Encoding the explainability measure ($\phi_{\explmeasure}$)}
The explainability of a decision diagram depends on the predicates used in it. To this end, we add the following conjunctions of soft clauses:
$$\bigwedge \limits_{1\leq i \leq k, p\in P} {\lambda'_{i,p}}$$
where:
$$\bigwedge \limits_{1\leq i \leq k, p\in P} {\lambda'_{i,p}} \leftrightarrow u_i \wedge \lambda_{i,p}$$
and define the weights of each of theses clauses $\lambda'_{i,p}$ based on the user-defined weights for using a predicate. 

Furthermore, the explainability will depend on the number of nodes used, and thus we will reward not using a node. To this end, we add the following conjunction of soft clauses:
$$\bigwedge \limits_{1\leq i\leq k} \overline{u_i}$$
where $u_i$ is true iff the node $i$ was used, i.e., is reachable from root node:
$$u_1 \wedge \bigwedge \limits_{2\leq i \leq k} u_i \leftrightarrow (\bigvee \limits_{\substack{1\leq c \leq c_{\max}, \\1\leq i'<i}} \tau_{i',c,i} \wedge u_{i'})$$

\paragraph{Encoding thresholds for \textsc{QuIntSynt}}
To restrict the space of interpretations to ones that have an explainability measure between two thresholds $\delta_{\interps}^l$ and $\delta_\interps^u$ we add an additional constraints that sums up the weights of satisfied soft clauses $\lambda_{i,p}$ and $\overline{u_i}$ and compares the result to $\delta_{\interps}^l$ and $\delta_\interps^u$. This is done by adding encoding for binary representations of the weights of each clause and encoding a binary adder that sums them up. 

\begin{enumerate}
	\item Encoding the weights: We assume that the weights are normalized to values between 0 and 1 that sum up to 1 and with floating precision 2, i.e., natural numbers representing the percentage between 0 and 100: 
		$$\bigwedge \limits_{1\leq i \leq k} (\overline{u_i} \rightarrow (w(u_i),b_{i,u},6) \wedge \overline{b^j_{i,\lambda}}) \wedge ({u_i} \rightarrow \bigwedge \limits_{0\leq j<7} \overline{b_{i,u}^j})$$
		and 
		$$\bigwedge \limits_{1\leq i \leq k, p\in P} ( \lambda'_{i,p} \rightarrow (w(\lambda_{i,p}),b_{i,\lambda},6))$$
	\item Encoding the adder:
		$$\bigwedge \limits_{0\leq i\leq 6} \overline{a_{0,u}^i} \wedge \overline{a_{0,\lambda}^i} $$
		and
		$$\bigwedge \limits_{1\leq i \leq k} \mathit{add}(a_{i+1,u},a_{i,u},b_{i,u},6)$$
		and 
		$$\bigwedge \limits_{1<\leq i \leq k} \mathit{add}(a_{i+1,\lambda},a_{i,\lambda},b_{i,\lambda},6)$$
		and 
		$$ \mathit{add}(a_\textit{fin}, a_{k+1,u},a_{k+1,\lambda},6 )$$
		where:
		 $$\mathit{add}(a_\gamma,a,b,k) =$$
		 $$\overline{c_\gamma^0}~ \wedge$$
		 $$\bigwedge \limits_{1\leq i\leq k} a_\gamma^i \leftrightarrow ((\overline{c_\gamma^i} \wedge (a^i \oplus b^i))\vee (c_\gamma^i \wedge (a^i \leftrightarrow b^i))) $$
		 $$\wedge \bigwedge \limits_{1< i \leq k} c_\gamma^i \leftrightarrow ((\overline{c_\gamma^{i-1}}\wedge a^{i-1}\wedge a^{i-1}) \vee (c_\gamma^{i-1}\wedge (a^{i-1}\vee b^{i-1}) ))$$
	\item Encoding the thresholds:
		$$ \delta_\interps^l \wedge \delta_\interps^u \wedge \mathit{smaller}(a_{\mathit{fin}}^0,\dots,a_{\mathit{fin}}^6,\delta_\interps^u) ~\wedge~ \mathit{larger}(a_{\mathit{fin}}^0,\dots,a_{\mathit{fin}}^6,\delta_\interps^l)$$
		where
		$$\mathit{smaller}(a_{\mathit{fin}}^0,\dots,a_{\mathit{fin}}^6,\delta)= $$
		$$\mathit{smaller^6} \wedge$$
		$$\bigwedge \limits_{1\leq i\leq 6} \mathit{smaller^i} \leftrightarrow ((\overline{a_{\mathit{fin}}^i} \wedge a_\delta^i) \vee (({a_{\mathit{fin}}^i} \leftrightarrow a_\delta^i) \wedge \mathit{smaller}^{i-1}))$$
		$$\wedge \mathit{smaller}^0 \leftrightarrow (\overline{a_{\mathit{fin}}^0} \wedge a_\delta^0) $$ 
		and 
		$$ \mathit{larger}(a_{\mathit{fin}}^0,\dots,a_{\mathit{fin}}^6,\delta)= $$
		$$\mathit{larger^6} \wedge$$
		$$\bigwedge \limits_{1\leq i\leq 6} \mathit{larger^i} \leftrightarrow (({a_{\mathit{fin}}^i} \wedge \overline{a_\delta^i}) \vee ( ({a_{\mathit{fin}}^i} \leftrightarrow a_\delta^i)\wedge \mathit{larger}^{i-1}) )$$
		$$\wedge \mathit{larger}^0 \leftrightarrow ({a_{\mathit{fin}}^0} \wedge \overline{a_\delta^0}) $$ 
\end{enumerate}

On feeding the above problem to a \textsc{MaxSAT} solver, it returns a satisfying assignment that gives a concrete instantiation of the interpretation template and maximizes $\corrmeasure$ and $\explmeasure$ in the interval $[\delta_\interps^l,\delta_\interps^u]$.

\subsection{Proofs of lemmas and theorems}
\subsubsection{Proofs from Section~\ref{sec:maxsat-enc}}
\begin{proof}[of Theorem~\ref{thm:pareto-opt} (Pareto-optimality)]
 	A solution $E$ for $\phi_{\langle \interps,\mathcal S, \corrmeasure, \explmeasure \rangle }$ with correctness and explainability measures $(c,e)$  is optimal with respect to $\corrmeasure + \explmeasure$, which in turn means that there is no interpretation $E'$ with measures $(c',e')$ such that $c'>c$ or $e'>e$. This implies that $E$ is Pareto-optimal with respect to $\corrmeasure$ and $\explmeasure$.
 \end{proof}

\subsubsection{Proofs from Section~\ref{sec:exploring-po-int}}
\begin{proof}[of Lemma~\ref{lem:sound} (Soundness)]
  We start by noting that from Theorem~\ref{thm:pareto-opt} we obtain that within a given interval \textsc{QuIntSynt} generates only Pareto-optimal interpretations. %

  Next we need to show that while iteratively using \textsc{QuIntSynt}, we output only Pareto-optimal points of the original problem instance. Indeed, the issue is that at each iterative call, since the interval changes (shrinks) we could get new Pareto-optimal points that are not Pareto-optimal in the original problem instance. We call such points as pseudo Pareto-optimal points. The soundness now follows by observing that a pseudo Pareto-optimal point occurs when a point that dominates was already Pareto-optimal and found in a previous iteration but has now been removed due to the shrinking of the interval. We can then show that this happens iff $c\leq \delta_{\mathcal C}$, which is precisely what is checked in Line 7 of  \Cref{alg:allintsynt} and hence such points are omitted.

To see this, we start by noting an invariant that holds at line 4 of \Cref{alg:allintsynt}: If $(E_l, E_u, c) = pop ( W )$, then there exists a Pareto-optimal point  $(c'', e'')$, where $c'' \ge c$ and $e'' \ge E_u$. This invariant can be proven inductively. The first time we arrive at line 4, this is true, because then $c = 0$ and $E_u = 1$. And we know that the most explainable interpretation has $(c'', e'') \ge (0, 1)$. Then, assuming that the invariant holds at line 4, we can show that everytime a $push(E_l', E_u', c')$ happens (to be popped later at some time in line 4), we also have the fact that there exists a Pareto-optimal point $(c'', e'')$ where $(c'' \ge c')$ and $(e'' \ge E_u')$. This follows from a straightforward case-analysis of pushes at line 9, 10 and 12.  

Now, it follows that if for the point $\interp$ found at line 5, $c\leq \delta_{\mathcal C}$, then by the invariant there is a point that dominates it and hence this point $\interp$ cannot be a true Pareto-optimal point. Conversely, if $c>\delta_{\mathcal C}$, we do not have any such restrictions and hence the algorithm proceeds. Thus, every point pushed in the algorithm is indeed a valid Pareto-optimal point, which proves the soundness of the algorithm. We also refer to the explanations on \Cref{fig:illustration_AllIntSynt} for more clarification.

 \end{proof}
  
\begin{proof}[of Lemma~\ref{lem:comp} (Completeness)]
This follows from (i) the fact that we have discrete and finitely many interpretations, i.e., range of $\explmeasure$, (ii) soundness which guarantees that every point computed by the algorithm is indeed pareto-optimal and (iii) monotonicity: every Pareto-optimal point continues to be Pareto-optimal after splitting the interval (i.e., across iterations). Essentially at each iteration, i.e., call to \textsc{QuIntSynt}, we get an interpretation with value $(c,e)$ and at line 9, 10 or 12, the interval reduces, which implies that the cardinality of range of $\explmeasure$ reduces by at least one. Hence the algorithm will terminate eventually from (i). But (iii) we know that each pareto-optimal point will be encountered in some iteration/sub-interval and by (ii) we are guaranteed that this point is output at that iteration.
\end{proof}

\begin{proof}[of Theorem~\ref{lem:wt} (Universality)]
The proof is in two steps. First if $\interp \in \arg\max \limits_{\interp'\in \interps} (\lambda(\corrmeasure(f_{\interp'},\mathcal S), \explmeasure(\interp')))$
then we claim that $\interp$ will be Pareto-optimal interpretation. To see this, we argue by contradiction. Suppose $\interp$ does not correspond to a Pareto-optimal point, then there exists $\interp'$ such that $\corrmeasure(f_{\interp'},\mathcal S)> \corrmeasure(f_{\interp},\mathcal S)$ and $\explmeasure(\interp')> \explmeasure(\interp)$. Since $\lambda$ is a strictly increasing function, this implies that $\lambda(\corrmeasure(f_{\interp'},\mathcal S), \explmeasure(\interp')))>\lambda(\corrmeasure(f_{\interp},\mathcal S), \explmeasure(\interp)))$ which is a contradiction as it violates the premise that $\interp \in \arg\max \limits_{\interp'\in \interps} (\lambda(\corrmeasure(f_{\interp'},\mathcal S), \explmeasure(\interp')))$.

Now, since $\interp$ is a Pareto-optimal interpretation, by Completeness Lemma~\ref{lem:comp}, our algorithm will find some interpretation with the same correctness and explainability measures as $\interp$, i.e., there exists ${\interp^\star}\in \interps$ such that $(\interp^\star, (\corrmeasure(f_{\interp^\star}, \mathcal S), \explmeasure(\interp^\star))) \in \textsc{ExplorePOI}(\interps, \mathcal S, \corrmeasure, \explmeasure)$ and (i) $\corrmeasure(f_\interp,\mathcal S) = \corrmeasure(f_{\interp^\star}, \mathcal S)$, (ii) $\explmeasure(\interp) = \explmeasure(\interp^\star)$.

\end{proof}

\subsection{Details on Benchmarks}
\paragraph{Decision module for predicting the performance of a perception module in an airplane (AP).}
The decision module predicts, based on the time of day, the cloud types, and the initial positioning of an airplane on a runway, whether a perception module used by the plane can be trusted to behave correctly. The decision module is an implementation of a decision tree that was trained on data collected from 200 simulations, using the XPlane\footnote{\url{x-plane.org}} simulator. The tree has more than 800 nodes. The labels in training data were determined based on whether the airplane exceeded a distance of 2.5m from the centerline for more than 10 computation steps. 

The input to the decision diagram is a tuple $(t,c,p)$ which defines the time of day, the cloud type, and the initial position of the plane on the runway. We defined the following three predicates for synthesis:
 \begin{itemize}
 	\item \emph{time of day}: this is a predicate defined by a feature function 
 	$$f_t\colon \mathbb R^3 \rightarrow \mathbb R, (t,c,p)\mapsto t$$ and a branching function
 	$$b_t\colon \mathbb R \rightarrow \{0,1,2\},t \mapsto \begin{cases}
 															 0 & 8am \leq t <12pm\\
 															 1 & 12pm \leq t < 6pm\\
 															 2 & otherwise \\	
 															\end{cases}$$
 	\item \emph{clouds}: this is a predicate defined by a feature function 
 	$$f_c\colon \mathbb R^3 \rightarrow \mathbb R, (t,c,p)\mapsto c$$ and a branching function
 	$$b_t\colon \{0,1,2,3,4,5\} \rightarrow \{0,1,2,3,4,5\},c \mapsto c$$
 	representing whether conditions with no clouds (branching 0) to dark clouds (branching 5).
 	\item \emph{initial position}: this is a predicate defined by a feature function 
 	$$f_p\colon \mathbb R^3 \rightarrow \mathbb R, (t,c,p)\mapsto p$$ and a branching function
 	$$b_p\colon \mathbb R \rightarrow \{0,1,2,3\},p \mapsto \begin{cases}
 															 0 & |p| < 0.5 \\
 															 1 & 0.5 \leq |p| < 2.5\\
 															 2 & 2.5 \leq  |p| < 3.5\\
 															 3 & otherwise 
 															\end{cases}$$
 	based on whether the plane is less than or more than 2.5 away from the centerline.
 \end{itemize}

Finally, two outputs were used in this benchmark, namely, \emph{alert} and \emph{no alert}.

\paragraph{Bank loan  predictor (BL).} 
The bank loan  predictor is a deep neural network that was trained on synthetic data that we created. The network was trained on the following features: age, monthly income, credit score, and the number of dependents. The training set included 100000 entries chosen such that the majority of people with age between 18 to 29 years, and those with age between 30 and 49 years but with income less than \$6000, were denied the loan. The values of the remaining features were chosen randomly. The network has five dense fully connected hidden layers with 200  ReLU's each, in addition to a Softmax layer and the output layer of two nodes.

The input to the decision diagram is a tuple $(a,i,c,d)$ which defines the age, income, credit score, and the number of dependents. We defined the following four predicates for synthesis:
 \begin{itemize}
 	\item \emph{age}: this is a predicate defined by a feature function 
 	$$f_a\colon \mathbb R^4 \rightarrow \mathbb R, (a,i,c,d)\mapsto a$$ and a branching function
 	$$b_a\colon \mathbb R \rightarrow \{0,1,2\},a \mapsto \begin{cases}
 															 0 & a <35\\
 															 1 & 35 \leq a < 60 \\
 															 2 & 60 	\leq a 		
 															\end{cases}$$
 	\item \emph{monthly income}: this is a predicate defined by a feature function 
 	$$f_i\colon \mathbb R^4 \rightarrow \mathbb R, (a,i,c,d)\mapsto i$$ and a branching function
 	$$b_a\colon \mathbb R \rightarrow \{0,1,2,3,4\},t \mapsto \begin{cases}
 															 0 & i < 2000\\
 															 1 & 2000 \leq i < 4000\\
 															 2 & 4000 \leq i < 6000 \\
 															 3 & 6000 	\leq i 		
 															\end{cases}$$
 	\item \emph{credit score}: this is a predicate defined by a feature function 
 	$$f_i\colon \mathbb R^4 \rightarrow \mathbb R, (a,i,c,d)\mapsto c$$ and a branching function
 	$$b_a\colon \mathbb R \rightarrow \{0,1\},c \mapsto \begin{cases}
 															 0 & c < 500\\
 															 1 & 500 \leq c 		
 															\end{cases}$$
 	\item \emph{dependents}: this is a predicate defined by a feature function 
 	$$f_i\colon \mathbb R^4 \rightarrow \mathbb R, (a,i,c,d)\mapsto d$$ and a branching function
 	$$b_a\colon \mathbb R \rightarrow \{0,1\},d \mapsto \begin{cases}
 															 0 & c < 3\\
 															 1 & 3 \leq c 		
 															\end{cases}$$
 \end{itemize}

Finally, two outputs were used in this benchmark, namely, \emph{approve} and \emph{deny}.

\paragraph{Theorem prover (TP).}
The neural network predicts the solvability of first-order formulas by a theorem prover with respect to percentage of unit clauses and the average clause length in a formula. The network had three hidden dense fully connected layers each with 200 ReLu's.  The data used to train the neural network can be found on the UCI machine learning repository under the following link \url{https://archive.ics.uci.edu/ml/datasets/First-order+theorem+proving}. The network was trained on the following features: F10, is a feature determining the average clause length in the formula, F1, is the percentage of unit clauses in the formula. For more details on the attributes we refer the reader to \cite{theoremProver}. 
 The authors of \cite{theoremProver} included data for five different heuristics H1-H5. We used the data for H1, thus predicting the solvability for H1. 
 
 The input to the decision diagram is a tuple $(f_1,f_{10})$ which defines the percentage of unit clause and the average clause length, respectively. We defined the following two predicates for synthesis:
 \begin{itemize}
 	\item \emph{F1}: this is a predicate defined by a feature function 
 	$$f_{F_1}\colon \mathbb R^4 \rightarrow \mathbb R, (f_1,f_{10})\mapsto f_1$$ and  branching functions
 	$$b_{F_1}\colon \mathbb R \rightarrow \{0,1,2,3\}, f_1 \mapsto \begin{cases}
 															 0 & f_1 <0.1\\
 															 1 & 0.1 \leq f_1 < 0.25 \\
 															 2 & 0.25 	\leq f_1 < 0.5\\ 
 															 3 & otherwise		
 															\end{cases}$$
 															
 	$$b_{F_1}\colon \mathbb R \rightarrow \{0,1\}, f_1 \mapsto \begin{cases}
 															 0 & f_1 <0.5\\
 															 1 & 0.25 \leq f_1 < 0.50 \\
 															 2 & otherwise		
 															\end{cases}$$
 	
 	$$b_{F_1}\colon \mathbb R \rightarrow \{0,1\}, f_1 \mapsto \begin{cases}
 															 0 & f_1 <0.5\\
 															 1 & otherwise		
 															\end{cases}$$
 															
 	\item \emph{F10}: this is a predicate defined by a feature function 
 	$$f_{F_{10}}\colon \mathbb R^4 \rightarrow \mathbb R, (f_1,f_{10})\mapsto f_{10}$$ and  branching functions
 	$$b_{F_{10}}\colon \mathbb R \rightarrow \{0,1\}, f_{10} \mapsto 															\begin{cases}
 															 0 & f_{10} <2\\
 															 1 & otherwise 	\end{cases}$$
 	$$b_{F_{10}}\colon \mathbb R \rightarrow \{0,1,2\}, f_{10} \mapsto 															\begin{cases}
 															 0 & f_{10} <2\\
 															 1 & 2 \leq f_{10} < 3 \\
 															 2 & 3	\leq f_{10}		\end{cases}$$
 	$$b_{F_{10}}\colon \mathbb R \rightarrow \{0,1,2,3\}, f_{10} \mapsto 															\begin{cases}
 															 0 & f_{10} <1\\
 															 1 & 1 \leq f_{10} < 2 \\
 															 2 & 2 \leq f_{10} < 3 \\
 															 3 & otherwise	\end{cases}$$
 \end{itemize}
 
 Finally, two outputs were used in this benchmark, namely, \emph{solvable} and \emph{not solvable}.

\subsection{Plotting executions times of explorations}
In \Cref{fig:plotAP,fig:plotBL,fig:plotTP} we plot the executions times of our the iterations of our exploration algorithm for all benchmarks on values $\delta=0.05$ and $\epsilon=0.05$. 
The diagrams show that for most Pareto-optimal points, 
the time taken to the find an interpretation was less than 20 seconds,
 but there were a few points in the Pareto-optimal 
 space where finding an interpretation took considerably more time.

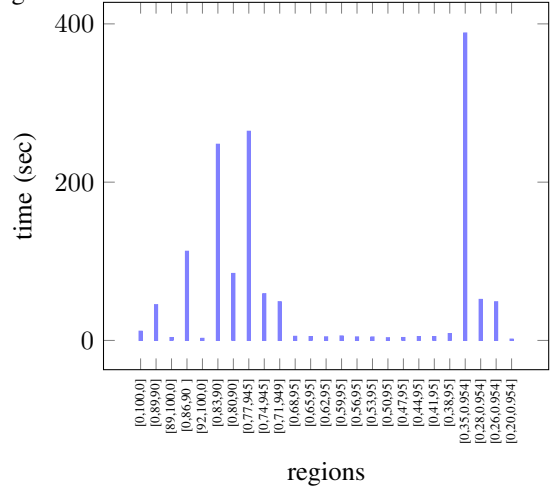
\begin{figure}[h!]
\begin{tikzpicture}
	\begin{axis}[xtick={0,1,2,3,4,5,6,7,8,9,10,11,12,13,14,15,16,17,18,19,20,21,22,23,24}, xticklabels = {{\tiny[0,100,0]},{\tiny[0,89,90]},{\tiny[89,100,0]},{\tiny[0,86,90 ]}, {\tiny[92,100,0]}, {\tiny[0,83,90]},{\tiny[0,80,90]}, {\tiny[0,77,945]}, {\tiny[0,74,945]},{\tiny[0,71,949]}, {\tiny[0,68,95]}, {\tiny[0,65,95]}, {\tiny[0,62,95]}, {\tiny[0,59,95]}, {\tiny[0,56,95]}, {\tiny[0,53,95]}, {\tiny[0,50,95]}, {\tiny[0,47,95]}, {\tiny[0,44,95]}, {\tiny[0,41,95]}, {\tiny[0,38,95]}, {\tiny[0,35,0.954]}, {\tiny[0,28,0.954]}, {\tiny[0,26,0.954]},{\tiny[0,20,0.954]} }, xticklabel style={rotate=90}, samples=50,xlabel={regions},
    ylabel={time (sec)}, every axis plot/.append style={
          ybar,
          bar width=.2,
          bar shift=0pt,
          fill
        }]

    \addplot[blue!50]coordinates {(0,11.687)};
    \addplot[blue!50]coordinates {(1,45.195)};
    \addplot[blue!50]coordinates {(2,3.677)};
    \addplot[blue!50]coordinates {(3,112.706)};
    \addplot[blue!50]coordinates {(4,2.665)};
    \addplot[blue!50]coordinates {(5,247.925)};
    \addplot[blue!50]coordinates {(6,84.648)};
    \addplot[blue!50]coordinates {(7,264.335)};
    \addplot[blue!50]coordinates {(8,59.085)};
    \addplot[blue!50]coordinates {(9,48.961)};
    \addplot[blue!50]coordinates {(10,5.206)};
    \addplot[blue!50]coordinates {(11,4.890)};
    \addplot[blue!50]coordinates {(12,4.561)};
    \addplot[blue!50]coordinates {(13,5.696)};
    \addplot[blue!50]coordinates {(14,4.475)};
    \addplot[blue!50]coordinates {(15,4.421)};
    \addplot[blue!50]coordinates {(16,3.4343)};
    \addplot[blue!50]coordinates {(17,3.701)};
    \addplot[blue!50]coordinates {(18,4.896)};
    \addplot[blue!50]coordinates {(19,4.888)};
    \addplot[blue!50]coordinates {(20,8.690)};
    \addplot[blue!50]coordinates {(21,388.526)};
    \addplot[blue!50]coordinates {(22,51.850)};
    \addplot[blue!50]coordinates {(23,48.881)};
    \addplot[blue!50]coordinates {(24,1.708)};

    \end{axis}
\end{tikzpicture}
\caption{Iteration runtimes: Airplane monitoring module}
\label{fig:plotAP}
\end{figure}

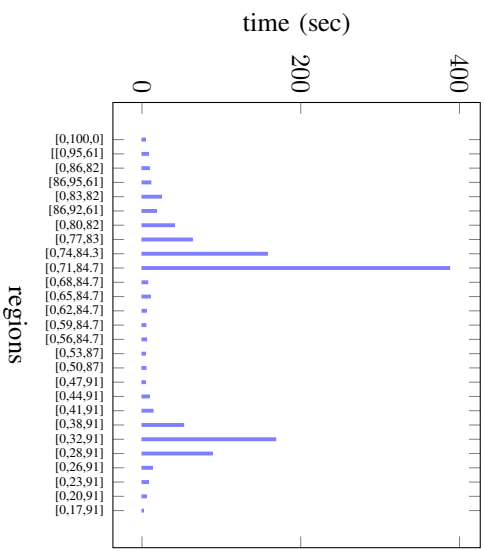
\begin{figure}[h!]
\begin{tikzpicture}
	\begin{axis}[xtick={0,1,2,3,4,5,6,7,8,9,10,11,12,13,14,15,16,17,18,19,20,21,22,23,24,25,26}, xticklabels = {{\tiny[0,100,0]},{\tiny[[0,95,61]},{\tiny[0,86,82]},{\tiny[86,95,61]}, {\tiny[0,83,82]}, {\tiny[86,92,61]},{\tiny[0,80,82]}, {\tiny[0,77,83]}, {\tiny[0,74,84.3]},{\tiny[0,71,84.7]}, {\tiny[0,68,84.7]}, {\tiny[0,65,84.7]}, {\tiny[0,62,84.7]}, {\tiny[0,59,84.7]}, {\tiny[0,56,84.7]}, {\tiny[0,53,87]}, {\tiny[0,50,87]}, {\tiny[0,47,91]}, {\tiny[0,44,91]}, {\tiny[0,41,91]}, {\tiny[0,38,91]}, {\tiny[0,32,91]}, {\tiny[0,28,91]}, {\tiny[0,26,91]},{\tiny[0,23,91]}, {\tiny[0,20,91]},{\tiny[0,17,91]} }, xticklabel style={rotate=90}, samples=50,xlabel={regions},
    ylabel={time (sec)}, every axis plot/.append style={
          ybar,
          bar width=.2,
          bar shift=0pt,
          fill
        }]

    \addplot[blue!50]coordinates {(0,4.272)};
    \addplot[blue!50]coordinates {(1,8.219)};
    \addplot[blue!50]coordinates {(2,9.001)};
    \addplot[blue!50]coordinates {(3,11.136)};
    \addplot[blue!50]coordinates {(4,24.591)};
    \addplot[blue!50]coordinates {(5,18.325)};
    \addplot[blue!50]coordinates {(6,40.968)};
    \addplot[blue!50]coordinates {(7,63.089)};
    \addplot[blue!50]coordinates {(8,157.909)};
    \addplot[blue!50]coordinates {(9,387.598)};
    \addplot[blue!50]coordinates {(10,7.195)};
    \addplot[blue!50]coordinates {(11,10.119)};
    \addplot[blue!50]coordinates {(12,5.160)};
    \addplot[blue!50]coordinates {(13,4.830)};
    \addplot[blue!50]coordinates {(14,5.787)};
    \addplot[blue!50]coordinates {(15,4.041)};
    \addplot[blue!50]coordinates {(16,4.989)};
    \addplot[blue!50]coordinates {(17,4.220)};
    \addplot[blue!50]coordinates {(18,8.947)};
    \addplot[blue!50]coordinates {(19,13.853)};
    \addplot[blue!50]coordinates {(20,52.511)};
    \addplot[blue!50]coordinates {(21,168.443)};
    \addplot[blue!50]coordinates {(22,88.430)};
    \addplot[blue!50]coordinates {(23,13.179)};
    \addplot[blue!50]coordinates {(24,8.122)};
    \addplot[blue!50]coordinates {(25,5.270)};
    \addplot[blue!50]coordinates {(26,1.927)};

    \end{axis}
\end{tikzpicture}
\caption{Iteration runtimes: Bank loan model}
\label{fig:plotBL}
\end{figure}

\begin{figure}[h!]
\begin{tikzpicture}
	\begin{axis}[ymax = 10, xtick={0,1,2,3,4,5,6,7,8,9,10,11,12,13,14,15,16,17,18,19}, xticklabels = {{\tiny[0,100,0]},{\tiny[0,96,82]},{\tiny[0,92,82]},{\tiny[0,88,86]}, {\tiny[0,84,86]}, {\tiny[0,80,86]},{\tiny[0,76,86]}, {\tiny[0,72,86]}, {\tiny[0,68,88]},{\tiny[0,64,0.88]}, {\tiny[0,60,90]}, {\tiny[0,56,90]}, {\tiny[0,52,90]}, {\tiny[0,48,90]}, {\tiny[0,44,90]}, {\tiny[0,40,90]}, {\tiny[0,36,90]}, {\tiny[0,32,90]}, {\tiny[0,28,0.90]}, {\tiny[0,24,90]} }, xticklabel style={rotate=90}, samples=50,xlabel={regions},
    ylabel={time (sec)}, every axis plot/.append style={
          ybar,
          bar width=.2,
          bar shift=0pt,
          fill
        }]

    \addplot[blue!50]coordinates {(0,0.766)};
    \addplot[blue!50]coordinates {(1,0.845)};
    \addplot[blue!50]coordinates {(2,0.957)};
    \addplot[blue!50]coordinates {(3,1.104)};
    \addplot[blue!50]coordinates {(4,1.586)};
    \addplot[blue!50]coordinates {(5,2.220)};
    \addplot[blue!50]coordinates {(6,3.391)};
    \addplot[blue!50]coordinates {(7,2.646)};
    \addplot[blue!50]coordinates {(8,1.075)};
    \addplot[blue!50]coordinates {(9,2.113)};
    \addplot[blue!50]coordinates {(10,1.177)};
    \addplot[blue!50]coordinates {(11,1.154)};
    \addplot[blue!50]coordinates {(12,1.143)};
    \addplot[blue!50]coordinates {(13,1.253)};
    \addplot[blue!50]coordinates {(14,1.131)};
    \addplot[blue!50]coordinates {(15,1.175)};
    \addplot[blue!50]coordinates {(16,1.105)};
    \addplot[blue!50]coordinates {(17,0.973)};
    \addplot[blue!50]coordinates {(18,0.992)};
    \addplot[blue!50]coordinates {(19,0.853)};

    \end{axis}
\end{tikzpicture}
\caption{Iteration runtimes: Theorem prover}
\label{fig:plotTP}
\end{figure}
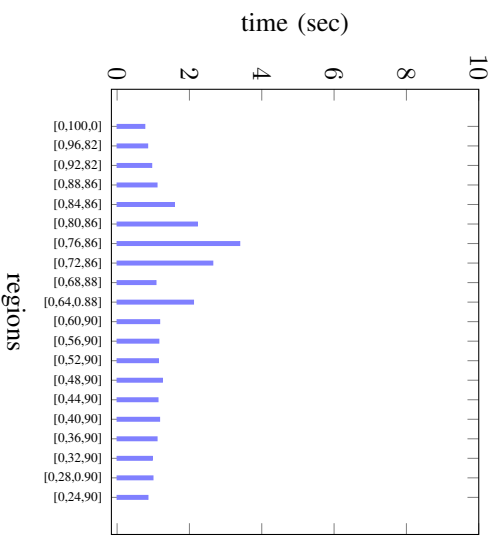

\end{document}